\pdfoutput=1

\documentclass{article}

\usepackage[nonatbib,final]{nips_2017}

\usepackage[utf8]{inputenc} 
\usepackage[T1]{fontenc}    
\usepackage{scrextend}
\usepackage{hyperref}       
\usepackage{url}            
\usepackage{booktabs}       
\usepackage{amsfonts}       
\usepackage{nicefrac}       
\usepackage{microtype}      
\usepackage{graphicx}
\usepackage{float}
\usepackage{subcaption}
\usepackage{multirow}
\usepackage{makecell}
\usepackage{footnote}

\usepackage{caption}
\makesavenoteenv{tabular}
\newcommand*\samethanks[1][\value{footnote}]{\footnotemark[#1]}

\newcommand{\specificthanks}[1]{\@fnsymbol{#1}}

\title{Reproducibility of Benchmarked Deep Reinforcement Learning Tasks for Continuous Control}

\author{
  Riashat Islam\thanks{Work done while interning at Maluuba, A Microsoft Company.}\hspace{.4em}\thanks{These two authors contributed equally.}\\
  School of Computer Science\\
  McGill University\\
  Montreal, QC, Canada\\
  \texttt{riashat.islam@mail.mcgill.ca}
  \And
  Peter Henderson\samethanks\\
  School of Computer Science\\
  McGill University\\
  Montreal, QC, Canada\\
  \texttt{peter.henderson@mail.mcgill.ca}
  \And
  Maziar Gomrokchi\\
  School of Computer Science\\
  McGill University\\
  Montreal, QC, Canada \\
  \texttt{maziar.gomrokchi@mail.mcgill.ca}\\
  \And
  Doina Precup\\
  School of Computer Science\\
  McGill University\\
  Montreal, QC, Canada \\
  \texttt{dprecup@cs.mcgill.ca}\\
}

\begin{document}

\maketitle

\begin{abstract}
Policy gradient methods in reinforcement learning have become increasingly prevalent for state-of-the-art performance in continuous control tasks. Novel methods typically benchmark against a few key algorithms such as deep deterministic policy gradients and trust region policy optimization. As such, it is important to present and use consistent baselines experiments. However, this can be difficult due to general variance in the algorithms, hyper-parameter tuning, and environment stochasticity. We investigate and discuss: the significance of hyper-parameters in policy gradients for continuous control, general variance in the algorithms, and reproducibility of reported results. We provide guidelines on reporting novel results as comparisons against baseline methods such that future researchers can make informed decisions when investigating novel methods.
\end{abstract}

\section{Introduction}

Deep reinforcement learning with neural network policies and value functions has had enormous success in recent years across a wide range of domains~\cite{DDPG,TRPO, TRPOGAE, DQN}. In particular, model-free reinforcement learning with policy gradient methods have been used to solve complex robotic control tasks~\cite{levine, Gu-Model-Based}. Policy gradient methods can be generally divided into two groups: off-policy gradient methods, such as Deep Deterministic Policy Gradients (DDPG)~\cite{DDPG} and on-policy methods, such as Trust Region Policy Optimization (TRPO)~\cite{TRPO}.


However, often there are many sources of possible instability and variance that can lead to difficulties with reproducing deep policy gradient methods. In this work, we investigate the sources of these difficulties with both on- and off-policy gradient methods for continuous control. We use two MuJoCo~\cite{MuJoCo} physics simulator tasks from OpenAI gym~\cite{Gym} (Hopper-v1 and Half-Cheetah-v1) for our experimental tasks. We investigate two policy gradient algorithms here: DDPG and TRPO. To our knowledge, there are few works~\cite{rllab} which reproduce existing policy gradients methods in reinforcement learning, yet many use as these algorithms as baselines to compare their novel work against~\cite{rllab,QPROP,SDQN,IPG}. We use the code provided by in~\cite{rllab} and~\cite{QPROP} for TRPO and DDPG (respectively), as these implementations are used in several works directly for comparison~\cite{IPG,QPROP,rllab,houthooft2016vime,rajeswaran2016epopt}


\textbf{Performance Measures : } We examine the general variance of the algorithms and address the importance of presenting all possible metrics across a large number of trials. Three main performance measures commonly used in the literature are: Maximum Average Return, Maximum Return, Standard Deviation of Returns, and Average Return. However, the first two measures are considered to be highly biased, while the last two are considered to be the most stable measures used to compare the performance of proposed algorithms. Thereby, in the rest of this work we only use the Average Return as our comparison measure unless stated otherwise, with final results displaying all metrics\footnote{We leave out Maximum Return unavenged across several trials, we posit this to be an unsuitable metric for reporting results. High-variance policies and environments may yield a vastly larger maximum return in an outlying trial.}.

\textbf{Hyper-parameter Settings : } We also highlight that there can be difficulty in properly fine-tuning hyper-parameter settings, leading to large variations of reported results across a wide range of works as different hyper-parameters are used. As in Tables~\ref{table:ddpg_papers} and~\ref{table:trpo_papers}, this inconsistency within the wide range of reported results makes it difficult to compare DDPG and TRPO as baseline algorithms without careful detailing of hyper-parameters, attention to the fairness of the comparison, and proper tuning of the parameters. Each cited work uses a different set of experimental hyper-parameters for supposed baseline comparisons\footnote{hyper-parameters for each paper can be found in detail in the references provided}. Running these algorithms with suboptimal hyper-parameter configurations may result in inaccurate comparisons against these baseline methods. As such, we highlight the significance of tuning various hyper-parameters and assess which of these yield the most significant differences in performance.

Based on our analysis, we encourage that careful consistency should be maintained when reporting results with both of these algorithms, as they are quite susceptible to hyper-parameters and the external sources of variance or randomness.

\section{Experimental Analysis}

We evaluate the off-policy DDPG~\cite{DDPG} and on-policy TRPO~\cite{TRPO} algorithms on continuous control environments from the OpenAI Gym benchmark~\cite{Gym}, using the MuJoCo physics simulator \cite{MuJoCo}. We empirically show the susceptibility and variance in results due to hyper-parameter configurations on two environments: Hopper ( $\mathcal{S} \subseteq \mathbb{R}^{20}, \mathcal{A} \subseteq \mathbb{R}^{3}$  ) and Half-Cheetah ($\mathcal{S} \subseteq \mathbb{R}^{20}, \mathcal{A} \subseteq \mathbb{R}^{6}$ ). All experiments\footnote{For code used, see: \url{https://github.com/Breakend/ReproducibilityInContinuousPolicyGradientMethods}.} are performed building upon the rllab Tensorflow implementation of TRPO~\cite{rllab} and the Q-Prop Tensorflow implementation of DDPG for our experiments~\cite{QPROP}.

\textbf{Experiment Details : }We run all variations for 5000 iterations and average all results across 5 runs. We investigate several hyper-parameters: batch size, policy network architecture, step size (TRPO), regularization coefficient (TRPO), generalized advantage estimation ($\lambda$) (TRPO), reward scale (DDPG), and actor-critic learning rates (DDPG). For each of these hyper-parameters we hold all others constant at default settings and vary the one under investigation across commonly used values. Lastly, we run a final set of experiments using the overall best cross-section of hyper-parameters for 10 trials using random seeds. We do this to investigate whether there is a significant difference in the results just due to variance caused by the random seeds.

For TRPO, the default hyper-parameters which we use are: a network architecture of (100,50,25) with ReLU hidden activations for a Gaussian Multilayer Perception Policy~\cite{rllab}; a step size of 0.01; a regularization coefficient of $1 \cdot 10^{-5}$; a Generalized Advantage Estimation $\lambda$ of 1.0~\cite{TRPOGAE}. For DDPG, we use default parameters as follows: a network architecture of ($100$,$50$,$25$) with relu hidden activations for a Gaussian Multilayer Perception Policy~\cite{rllab}; actor-critic learning rates of $1\cdot10^{-3}$ and $1\cdot10^{-4}$; batch sizes of $64$; and a reward scale of $0.1$.

\subsection{Common Hyper-Parameters}

First, we investigate several hyper-parameters common to both TRPO and DDPG: policy architecture and batch size. We use the same sets of hyper-parameters as reported in previous works using these implementations in an attempt to reproduce the results reported in these works.



\textbf{{Policy Network Architecture : }} The policy network architecture can play an important role in the maximum reward achieved by the algorithm due to the amount of information storage provided by the network. We use a hidden layer sizes ($64$,$64$) as in~\cite{TRPO}, ($400$,$300$) as in~\cite{rllab,DDPG}, and ($100$,$50$,$25$) as in~\cite{QPROP} for comparing the results of these algorithms\footnote{All of these use RELU activations for the hidden layers and a Gaussian MLP Policy.}.

\begin{figure}[ht!]
  \includegraphics[width=.5\textwidth]{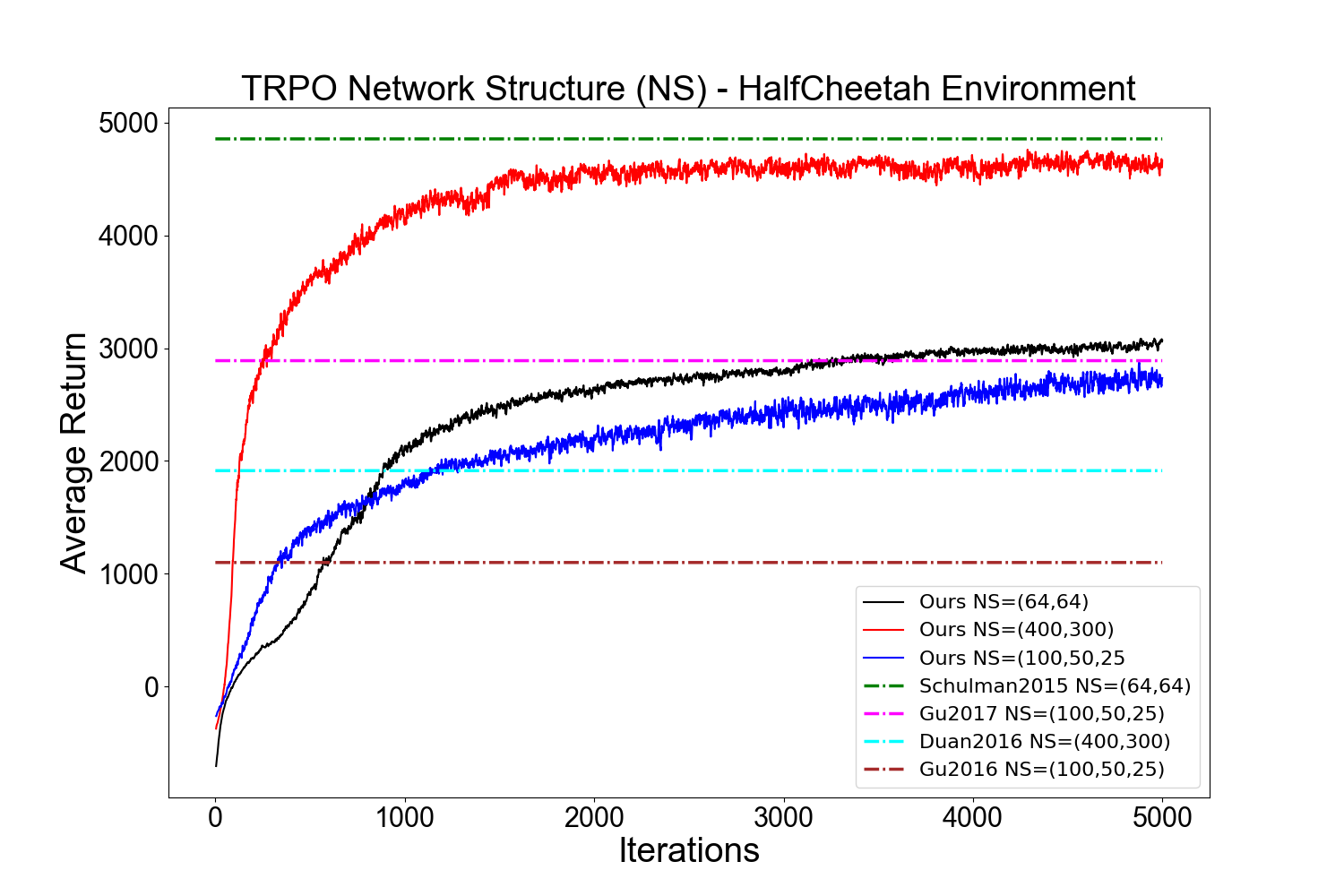}
  \includegraphics[width=.5\textwidth]{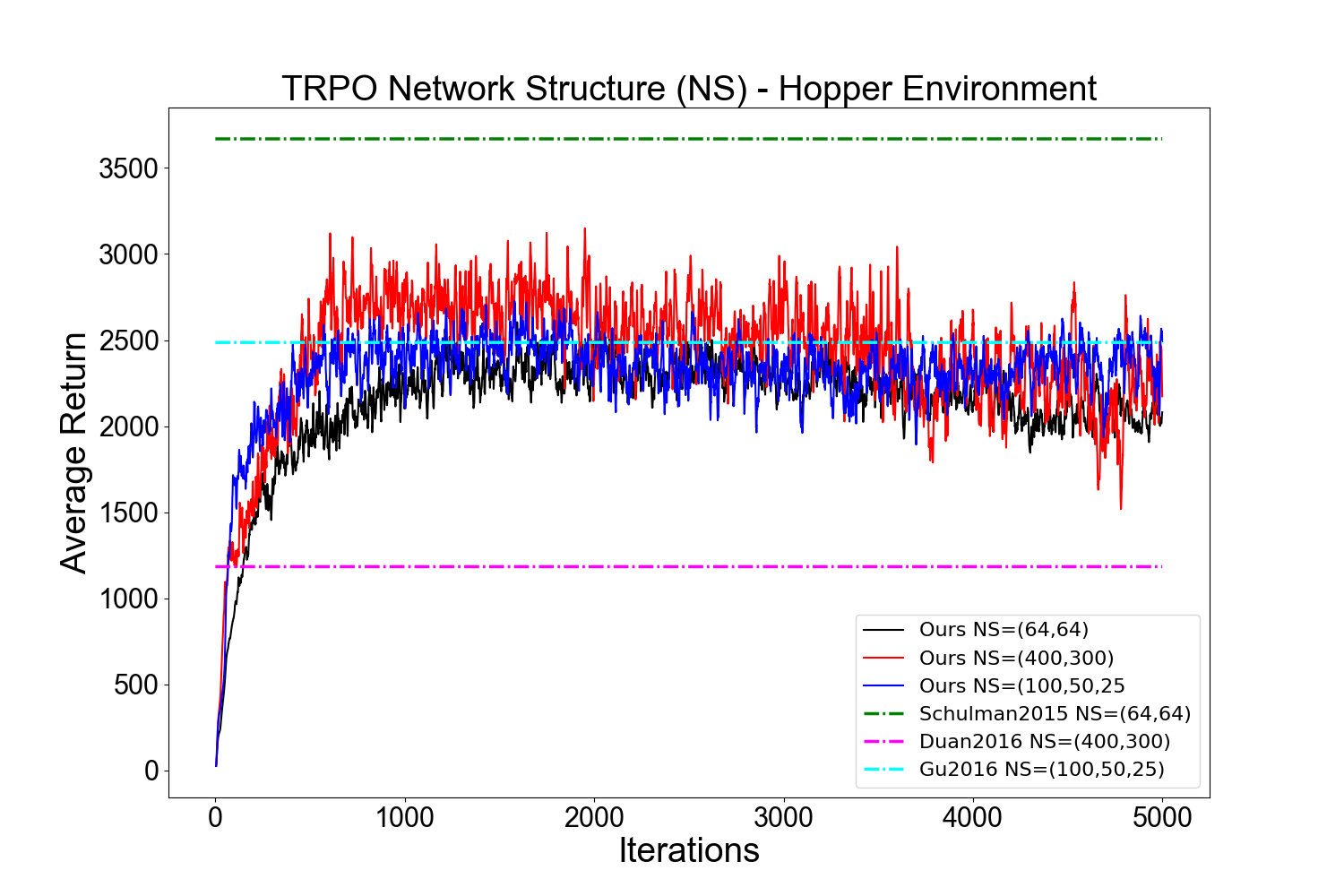}
\caption{TRPO on Half-Cheetah with different network configurations}
\label{trpo_arch}
\end{figure}

\begin{figure}[ht!]
\includegraphics[width=.5\textwidth]{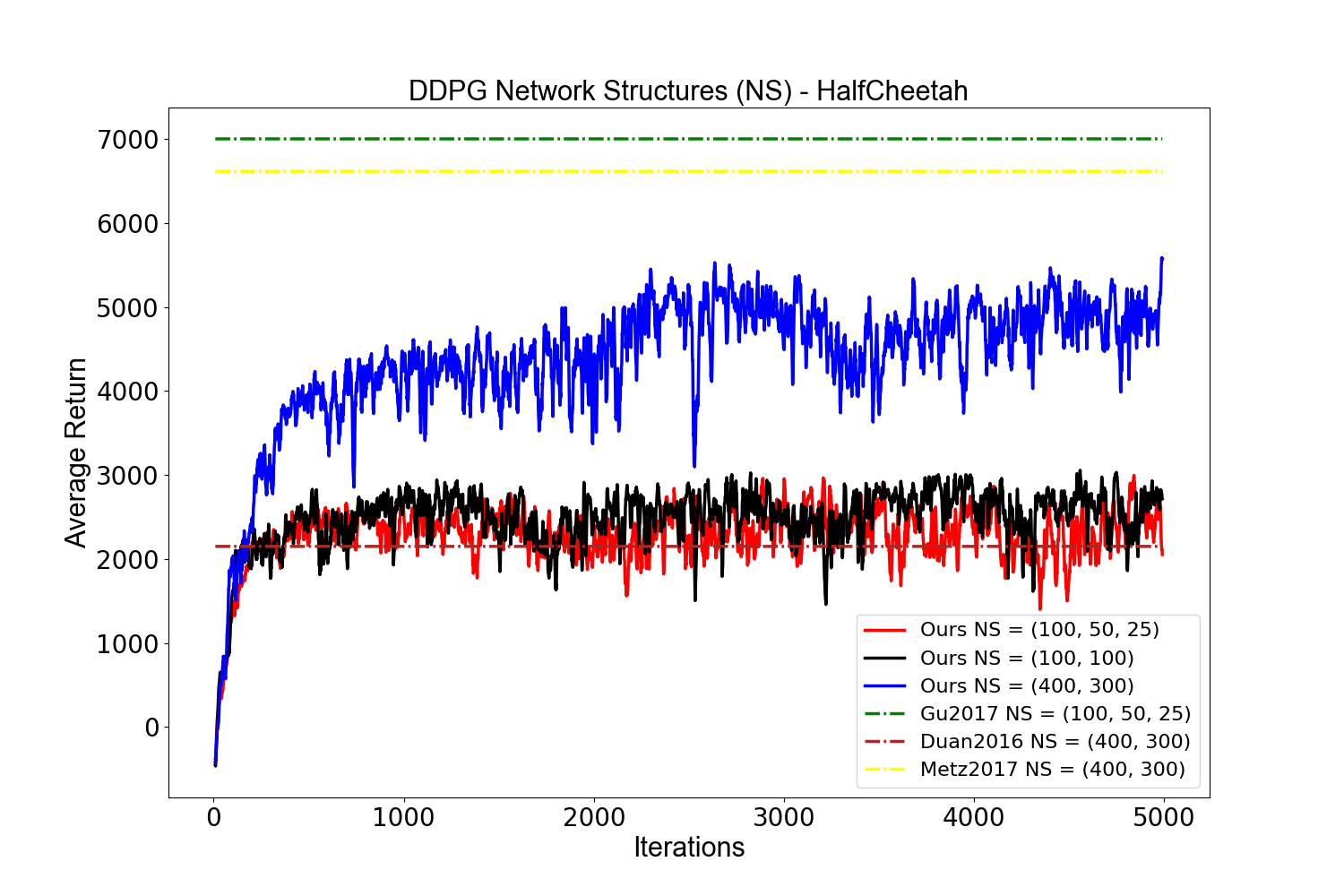}
\includegraphics[width=.5\textwidth]{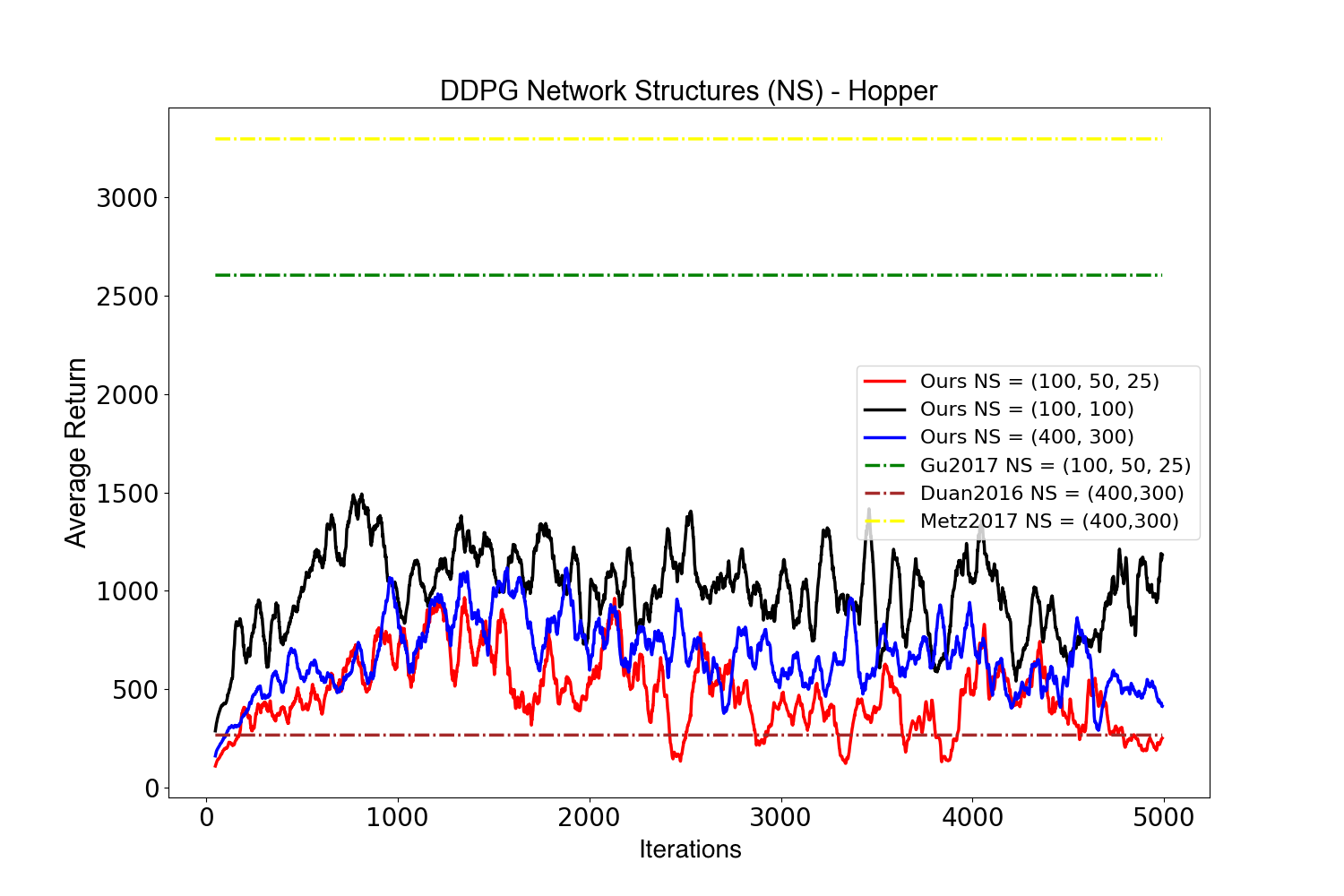}
\caption{DDPG on Half-Cheetah and Hopper on different network configurations}
\label{ddpg_network}
\end{figure}

Our results can be found in Figures~\ref{trpo_arch} and~\ref{ddpg_network}. Notably, the ($400$,$300$) architecture significantly outperforms both other smaller architectures for Half-Cheetah and to a less significant extent Hopper as well\footnote{For TRPO Half-Cheetah using a two-sample t-test on the sample rollouts: against ($64$,$64$) $t=-13.4165,p=0.0000$; against ($100$,$50$,$25$) $t=-11.3368,p=0.0016$. For TRPO Hopper: against ($100$,$50$,$25$) $t=-0.5904,p=0.2952$; against ($64$,$64$) $t=-1.9081,p=0.2198$}. This is true for both TRPO and DDPG. However, the architecture which we found to be the best ($400$,$300$) is not the one which is used in reporting results for baselines results in~\cite{QPROP,IPG}.

For the Hopper environment, for both TRPO and DDPG, results are not as significantly impacted by varying the network architecture, unlike the Half-Cheetah environment. This is somewhat thematic of what we find across all hyper-parameter variations on Hopper, as will be further discussed later. In particular, our investigation of DDPG on different network configurations shows that for the Hopper environment, DDPG is quite unstable no matter the network architecture. This can be attributed partially to the high variance of DDPG itself, but also to the increased stochasticity of the Hopper task. As can be seen in Figure~\ref{ddpg_network}, even with varied network architectures, it is difficult to tune DDPG to reproduce results from other works even when using their reported hyper-parameter settings.

\textbf{Batch Size : }The batch size parameter plays an important role in both DDPG and TRPO. In the off-policy DDPG algorithm, the actor and critic updates are made by sampling a mini-batch uniformly from the replay buffer. Typically, the replay buffer is allowed to be large. In~\cite{DDPG} and~\cite{QPROP}, a batch size of $64$ was used, whereas the original rllab implementation uses a batch size of $32$. Our analysis with different mini-batches for DDPG $(32, 64, 128)$ shows that similar performance can be obtained with mini-batch sizes of $32$ and $64$, whereas significant improvements can be obtained with a batch size of $128$.

\begin{figure}[ht!]
\includegraphics[width=.5\textwidth]{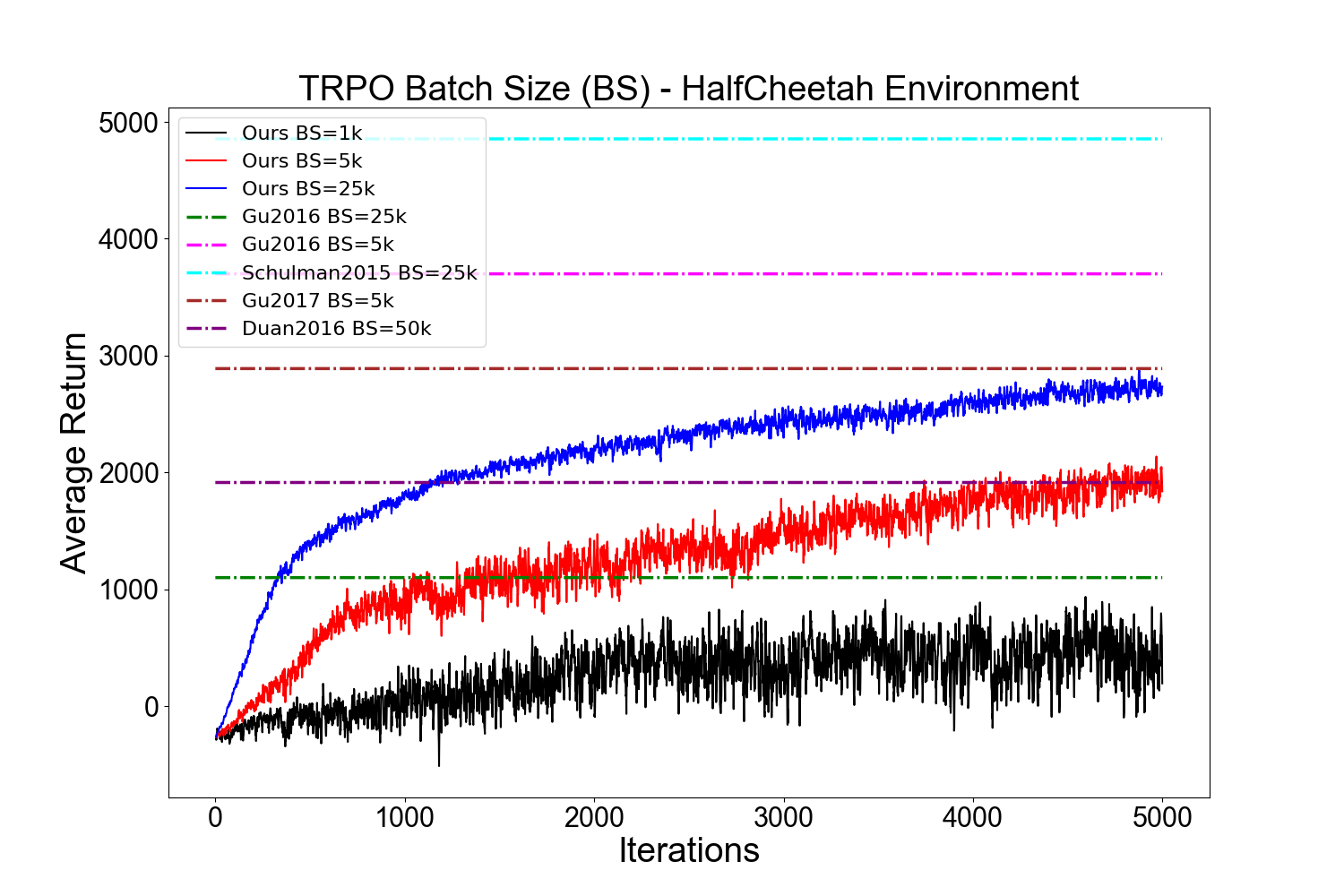}
\includegraphics[width=.5\textwidth]{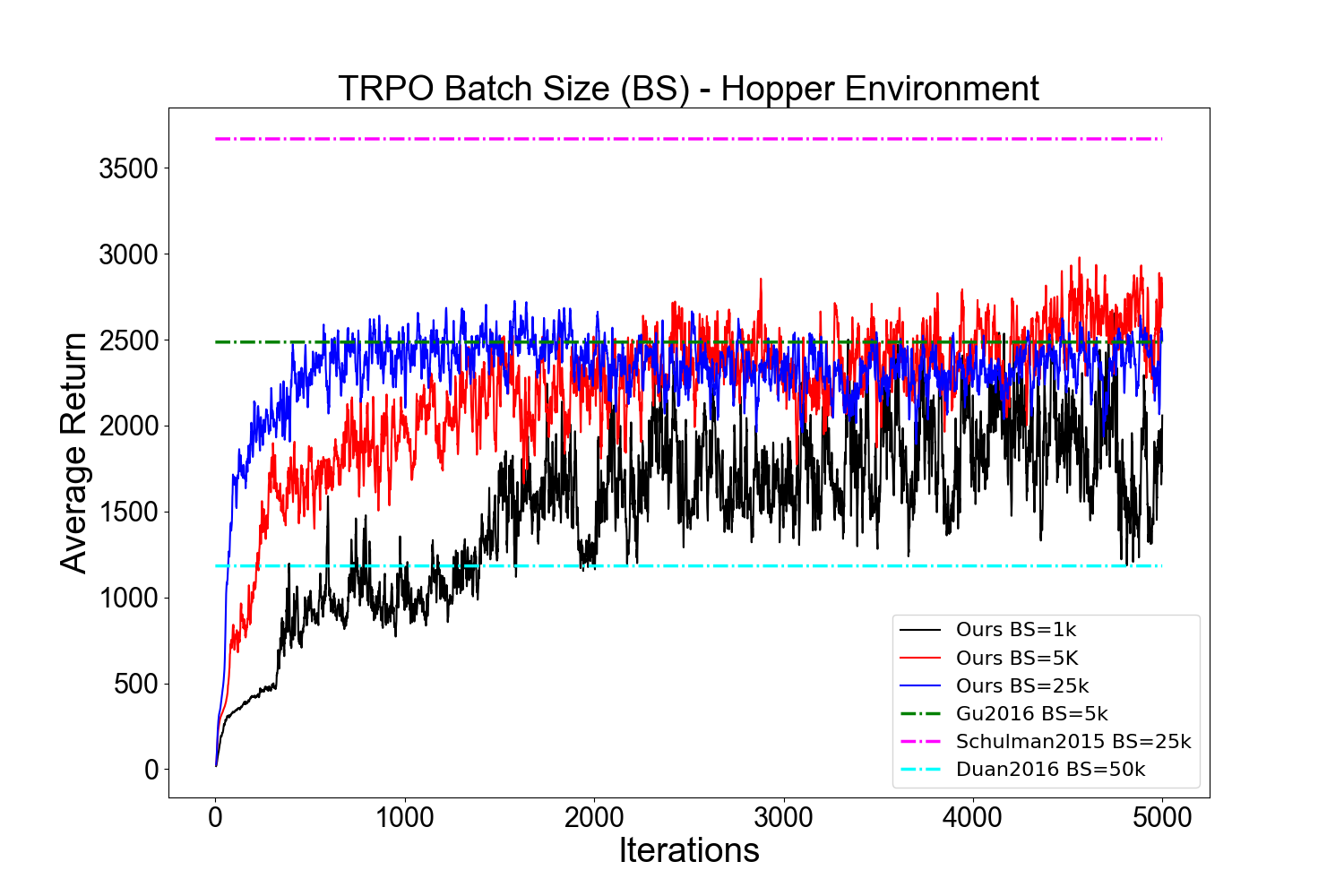} 
\caption{TRPO on Half-Cheetah and Hopper - Significance of batch size}
\label{trpo_batch}
\end{figure}

For TRPO, larger batch sizes are necessary in general. We investigate the same batch sizes as used in~\cite{QPROP,TRPO} of ($1000$,$5000$,$25000$). As expected, a batch size of $25000$ produces the best results. As we constrain learning to $5000$ episodes, it is intuitive that a larger batch size would perform better in this time frame as more samples are seen. Furthermore, as can be seen in Figure~\ref{trpo_batch} for Half-Cheetah, the smaller batch sizes begin to plateau to a much lower optimum.


By intuition, this may be due to TRPO's use of conjugate gradient optimization with a KL constraint. With small sample batch sizes, gradients differences between steps may be much larger in a high variance environment and results in a more unstable training regime.


\begin{figure}[ht!]
\includegraphics[width=.5\textwidth]{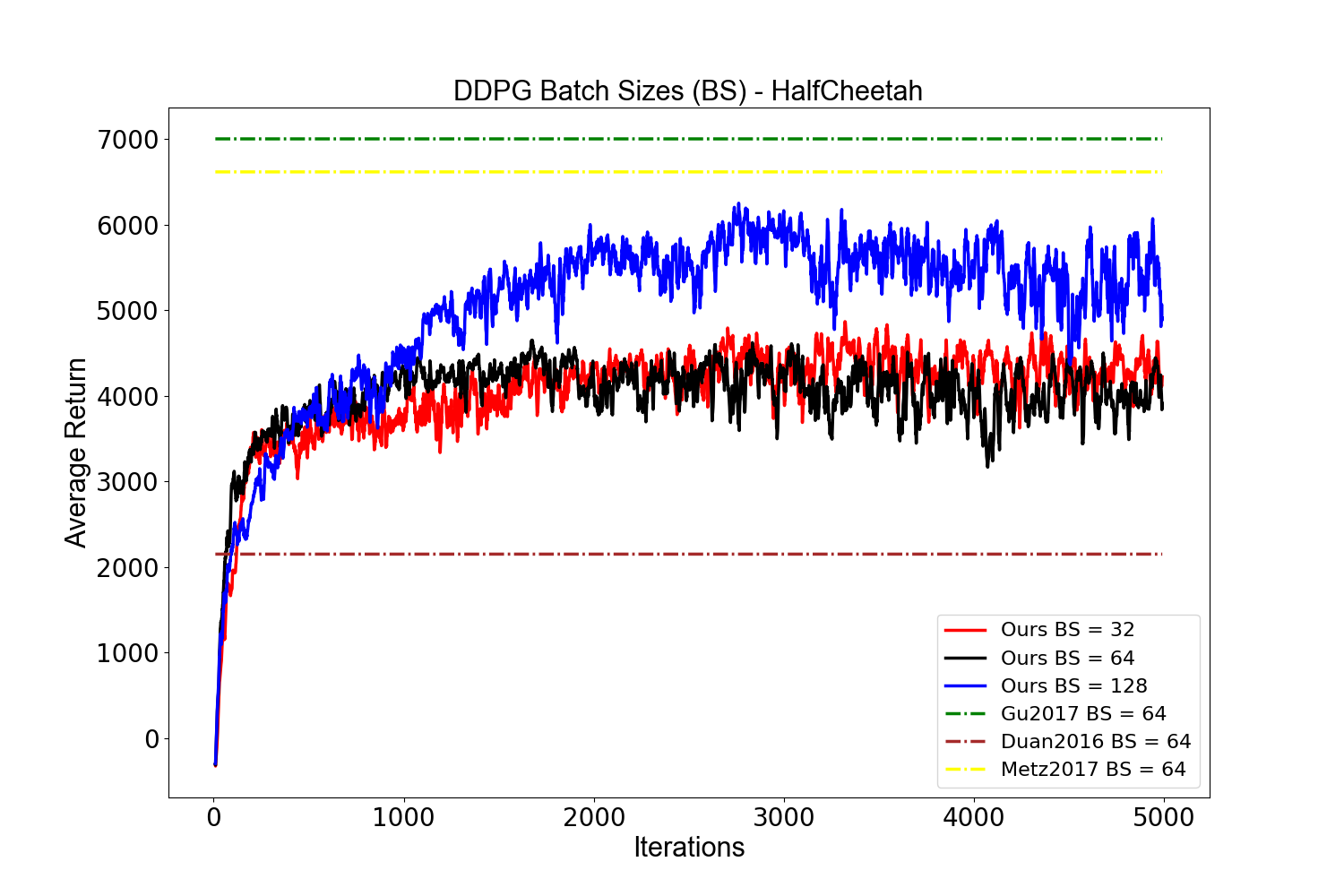}
\includegraphics[width=.5\textwidth]{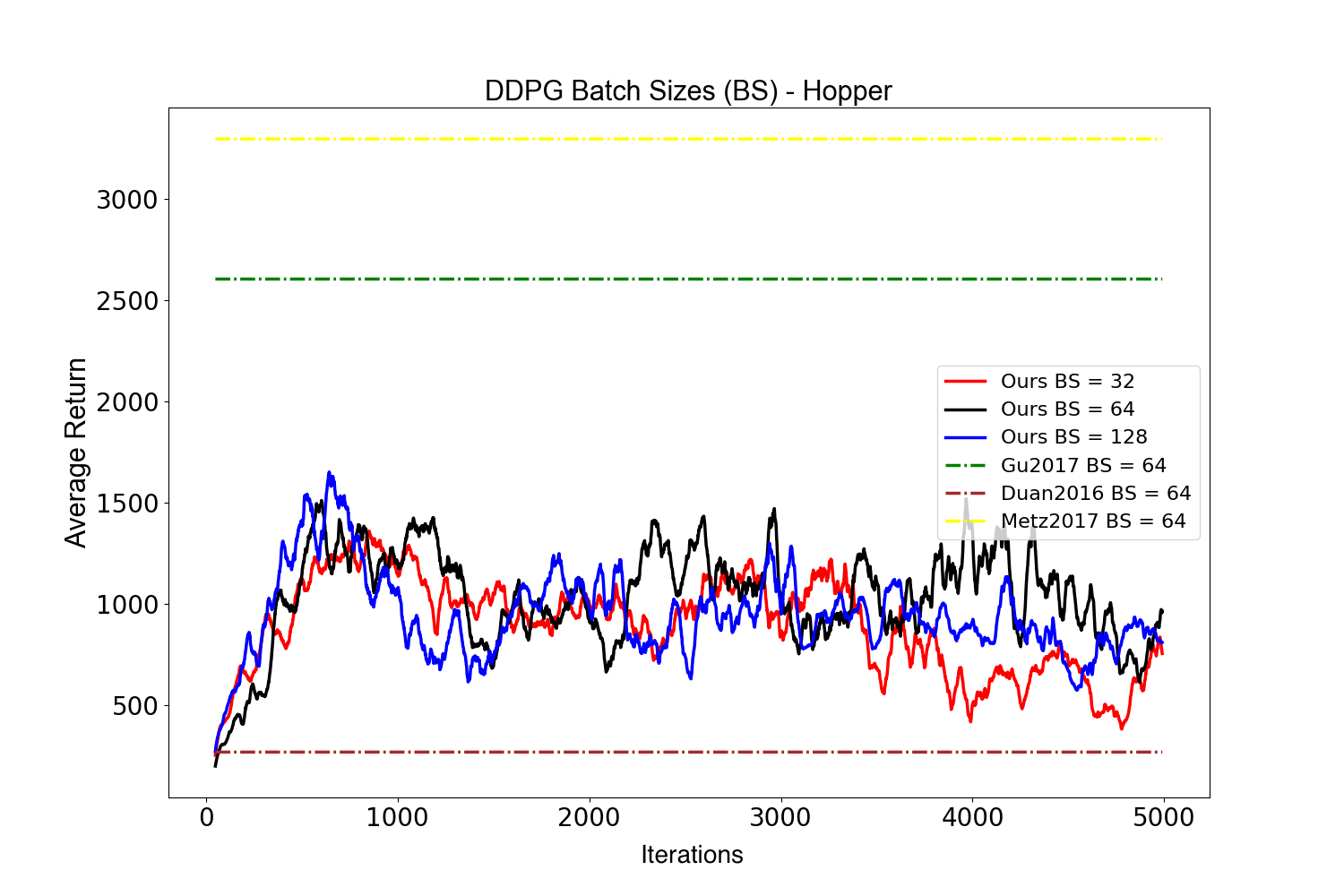}
\caption{DDPG on Half-Cheetah and Hopper - Significance of the mini batch size}
\label{ddpg_batch}
\end{figure}

We also highlight that the DDPG algorithm with different batch sizes produces similar results for the Hopper environment. While other works have reported different tuned parameters for DDPG, we establish the high variance of this algorithm, producing similar results with different batch sizes for the Hopper environment, while a larger batch size improves performance in Half-Cheetah as seen in Figure~\ref{ddpg_batch}.

\subsection{TRPO-Specific Hyper-Parameters}

\begin{figure}[ht!]
\includegraphics[width=.5\textwidth]{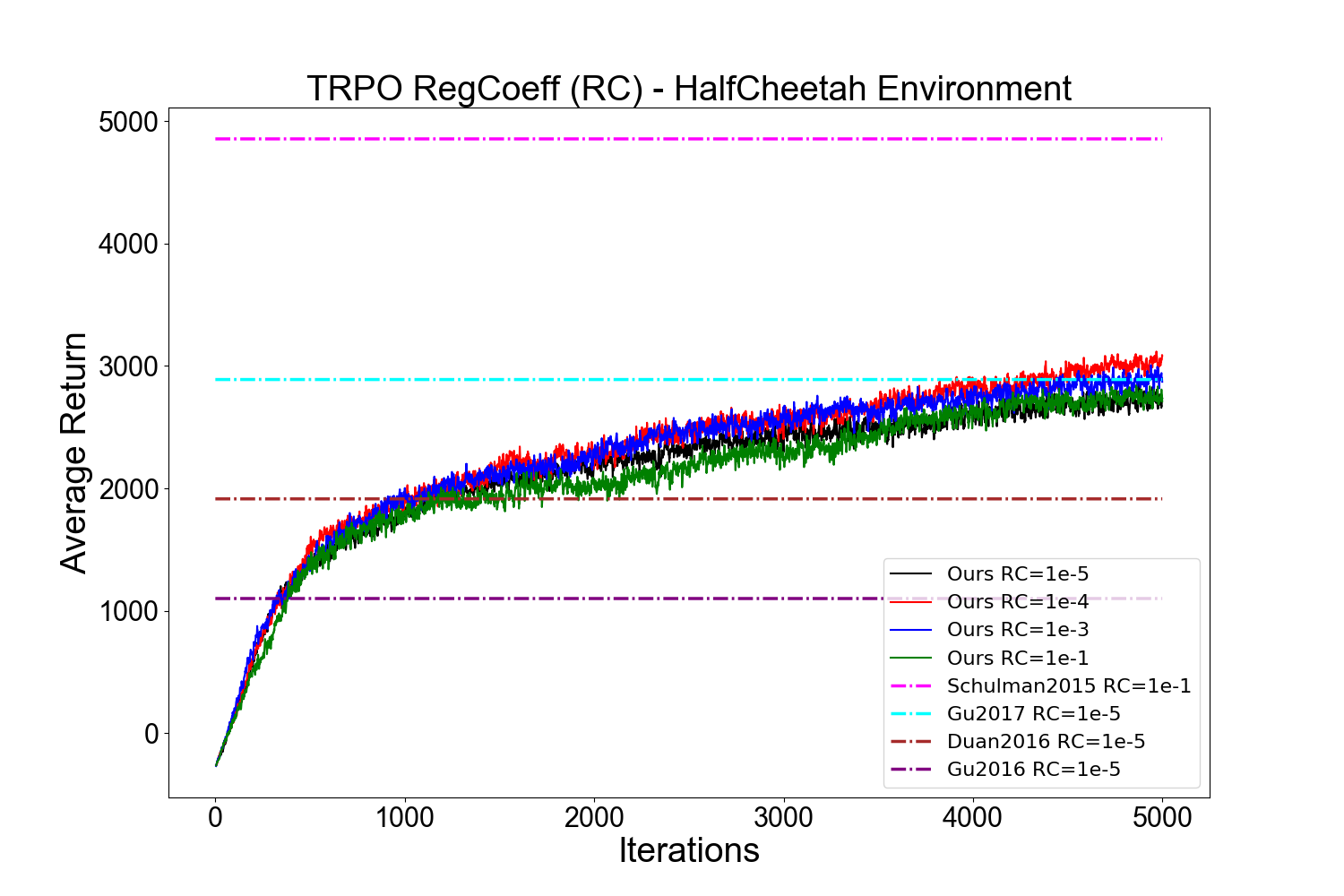}
\includegraphics[width=.5\textwidth]{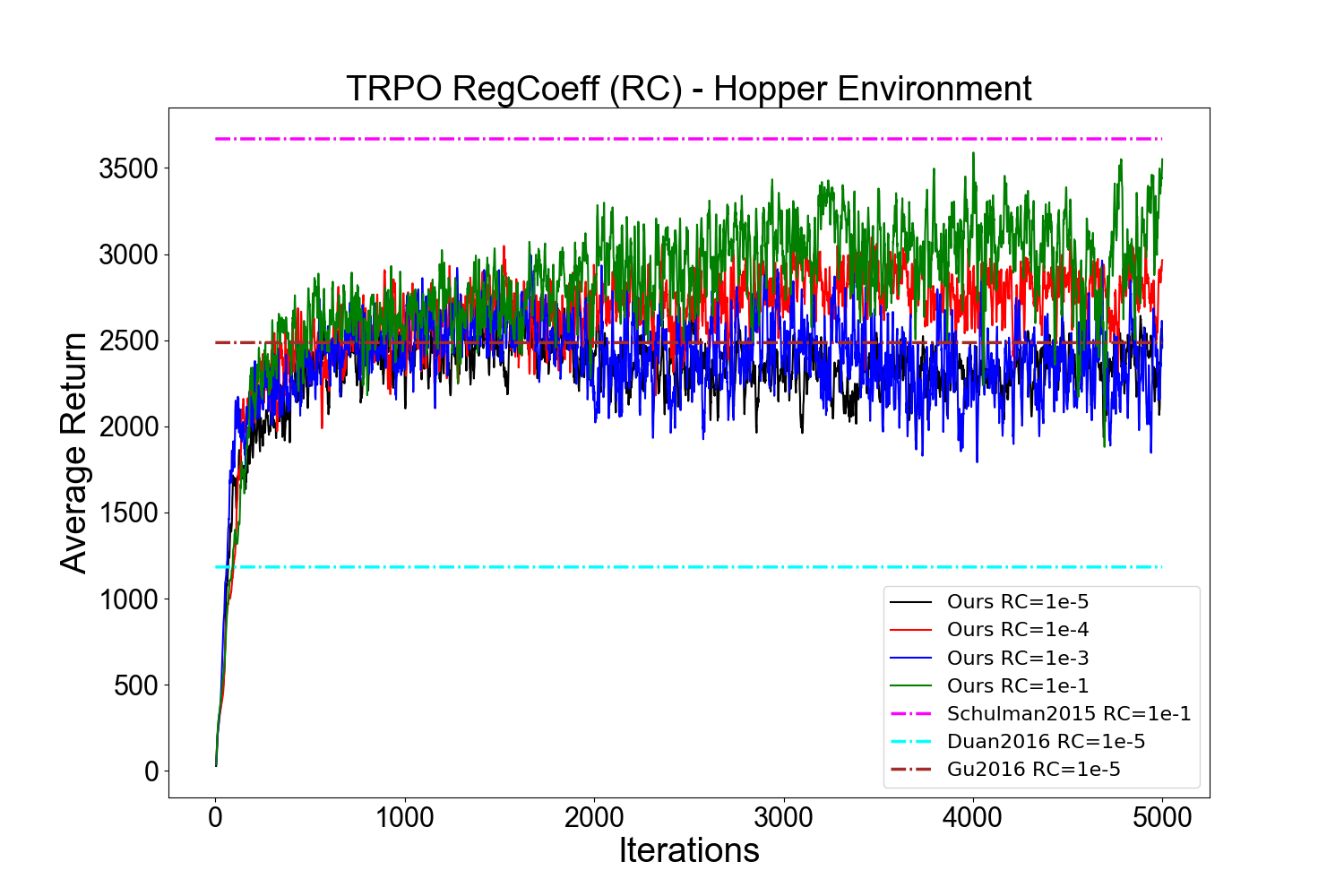}
\caption{Regularization coefficient variations for TRPO. Cited implementation values may use different sets hyper-parameters. See associated works for specific details.}
\label{trpo_reg}
\end{figure}

\textbf{{Regularization Coefficient} : }The regularization coefficient (RC) (or conjugate gradient damping factor) is used as a regularizer by adding a factor of the identity matrix to the Fisher matrix (or finite difference HVP in~\cite{rllab}) during the conjugate gradient step. We investigate a range of values between $1\cdot10^{-5}$ to $.1$ based on values used in aforementioned works. We don't see a significant difference\footnote{Using an average of 2-sample t-test comparisons, the largest difference from the default parameter in Hopper is $t=2.8965,p=0.1443$ with RC=0.1 and $t=0.8020, p=0.4540$ with RC=.0001.} in using one particular value of RC over another, though it seems to have a more significant effect on Hopper. Figure~\ref{trpo_reg} shows the average learning graphs for these variations.

\begin{figure}[ht!]
\includegraphics[width=.5\textwidth]{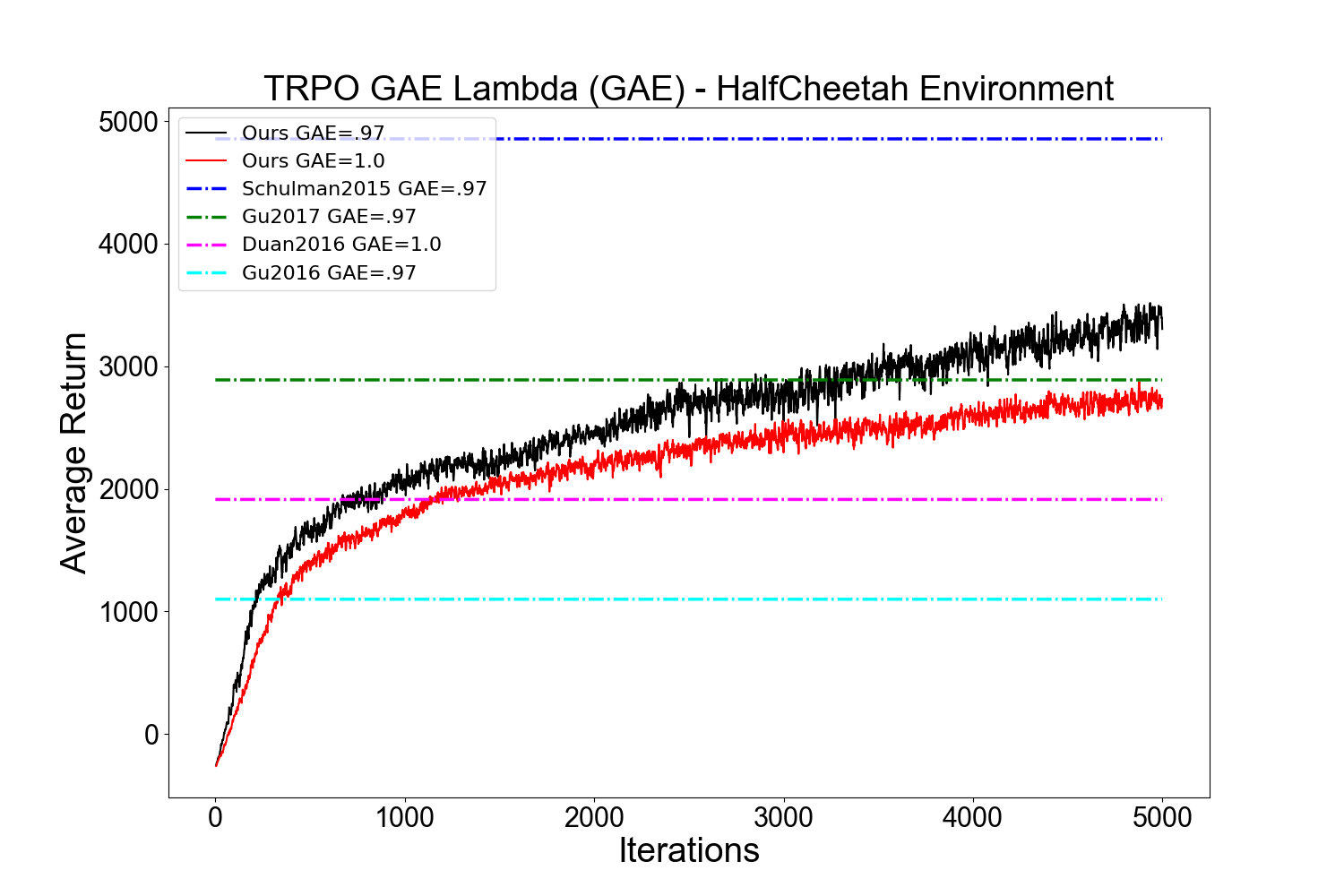}
\includegraphics[width=.5\textwidth]{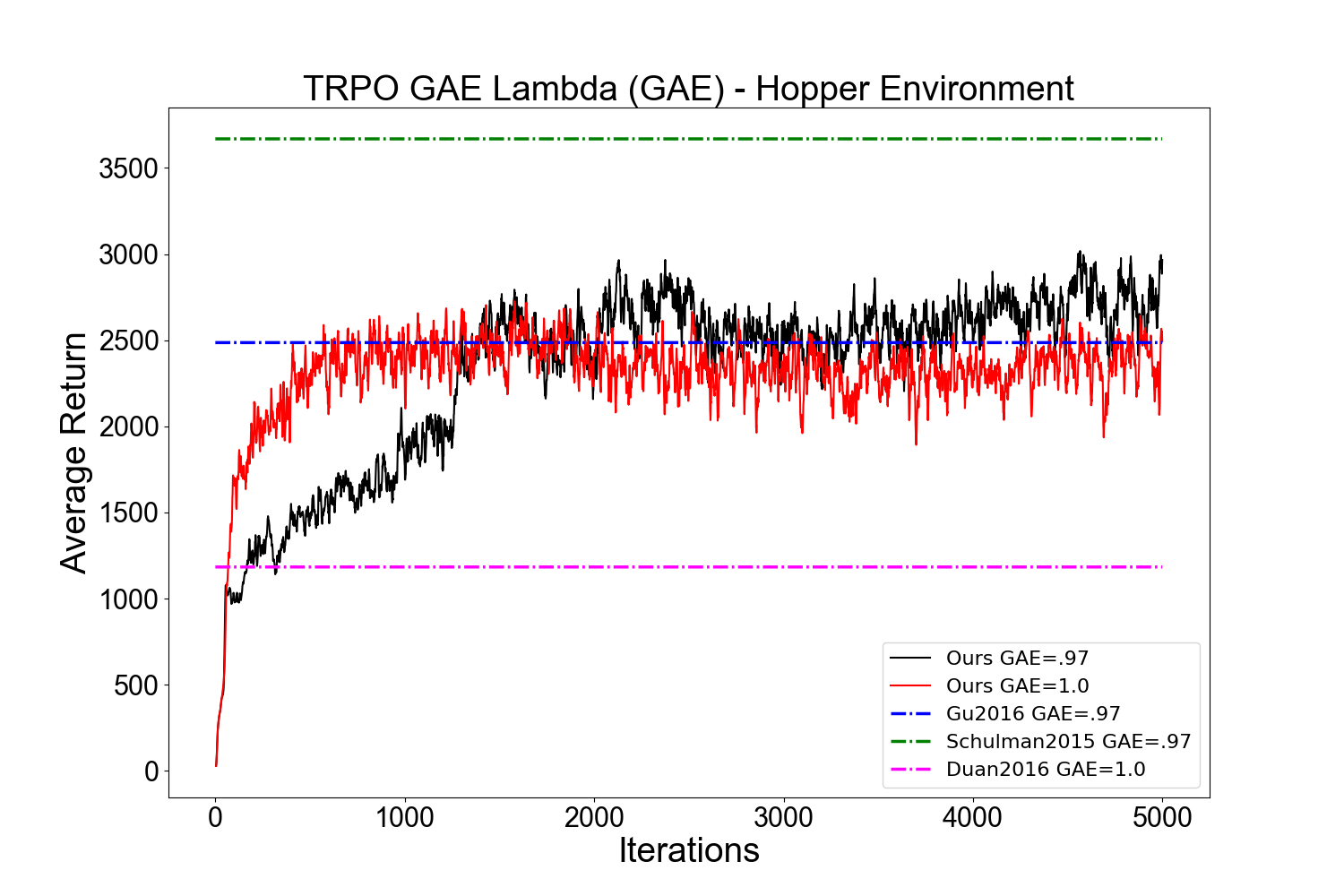}
\caption{Generalized advantage estimation lambda value variations for TRPO. Cited implementation values may use different sets hyper-parameters. See associated works for specific details.}
\label{trpo_gae}
\end{figure}

\textbf{{Generalized Advantage Estimation : }} Generalized advantage estimation~\cite{TRPOGAE} has been shown to improve results dramatically for TRPO. Here, we investigate using $\lambda=1.0$ and $\lambda=.97$ for this. We find that for longer iterations, a lower GAE $\lambda$ does in fact improve results for longer sequences in Half-Cheetah and mildly for Hopper\footnote{For Half-Cheetah, $t=2.9109,p=0.0652$ for last 500 iterations and $t=1.9231,p=0.1978$ overall. For Hopper, $t=1.9772,p=0.1741$ for last 500 iterations and $t=-0.1255,p=0.2292$ overall.}. Figure~\ref{trpo_gae} shows the average learning graphs for these variations.

\begin{figure}[ht!]
  \includegraphics[width=.5\textwidth]{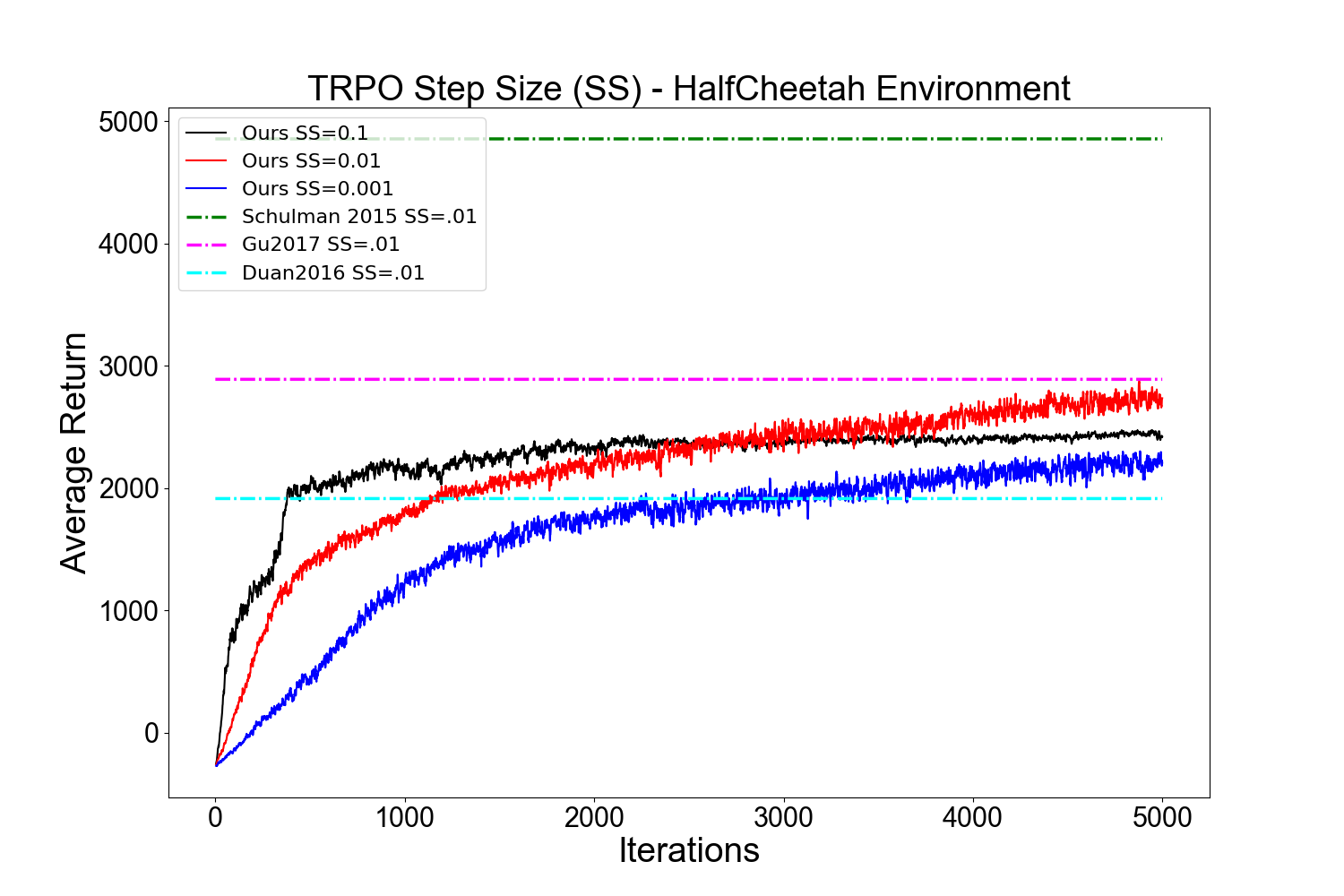}
  \includegraphics[width=.5\textwidth]{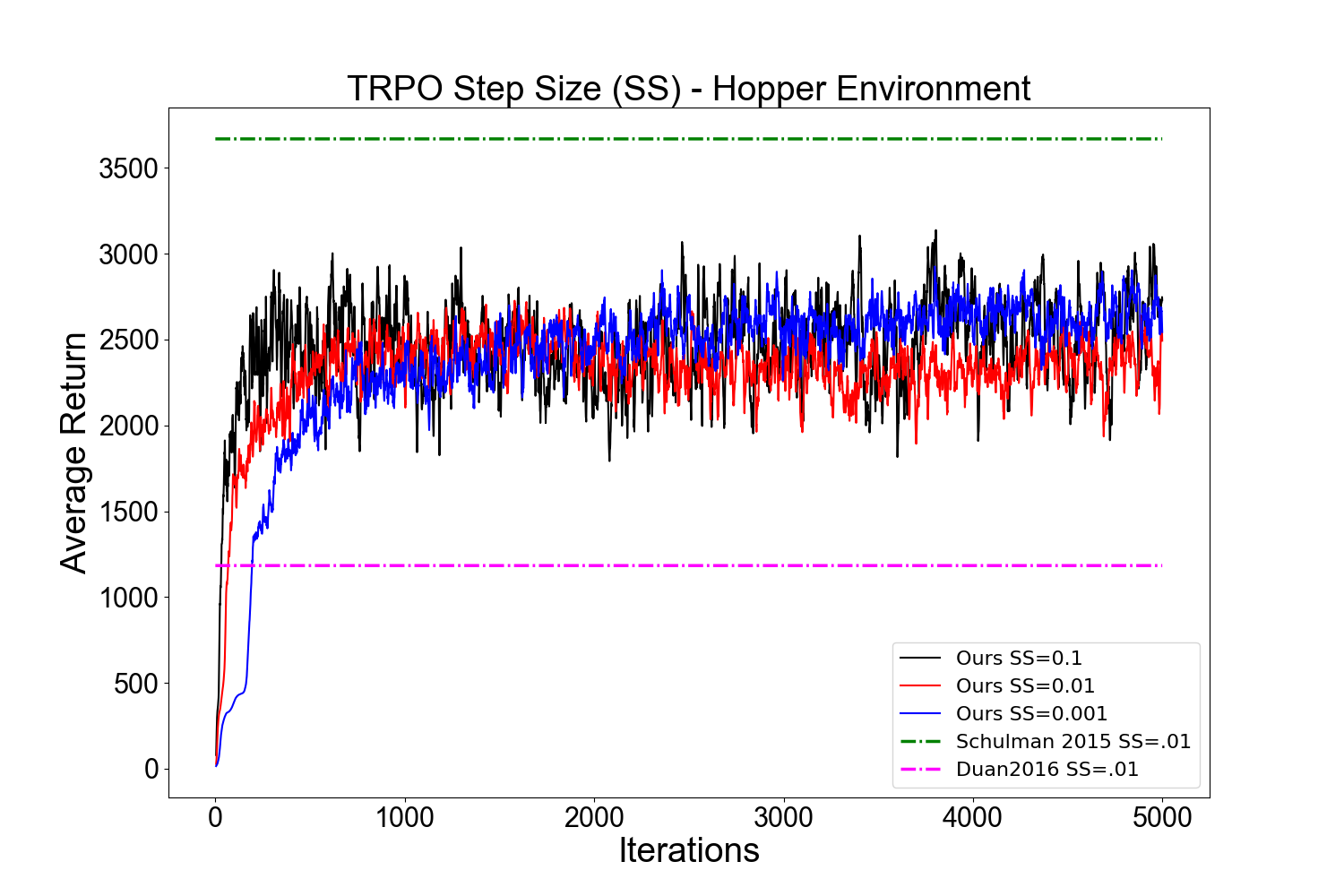}
\caption{Step size variations for TRPO. Cited implementation values may use different sets hyper-parameters. See associated works for specific details.}
\label{trpo_step}
\end{figure}

\textbf{{Step Size : }}The step size (SS) (effectively the learning rate of TRPO) is the same as the KL-divergence bound for the conjugate gradient steps. Here, we find that the default value of 0.01 appears to work generally the best for both Hopper and Half-Cheetah\footnote{Hopper most significant t-test difference from default is SS=0.1 with $t=1.0302,p=0.2929$, and for Half-Cheetah difference from default and SS=0.001 $t=-3.1255,p=0.0404$}.  Figure~\ref{trpo_step} shows the average learning curves for these variations. The intuition here is the same behind adjusting learning rates in standard gradient optimization methods, though the formulation is through a constraint rather than a learning rate, it effectively has the same characteristics when tuning it.

\subsection{DDPG-Specific Hyper-Parameters}

We investigate two hyper-parameters which are unique to DDPG which previous works have described as important for improving results~\cite{rllab,QPROP}: reward scale and actor-critic learning rates.

\textbf{Reward Scale : }As in \cite{rllab}, all the rewards for all tasks were rescaled by a factor of $0.1$ to improve the stability of DDPG. It has been claimed that this external hyper-parameter, depending on the task, can make the DDPG algorithm unstable. Experimental results in \cite{QPROP} give indication that DDPG is particularly sensitive to different reward scale settings.

\begin{figure}[ht!]
\includegraphics[width=.5\textwidth]{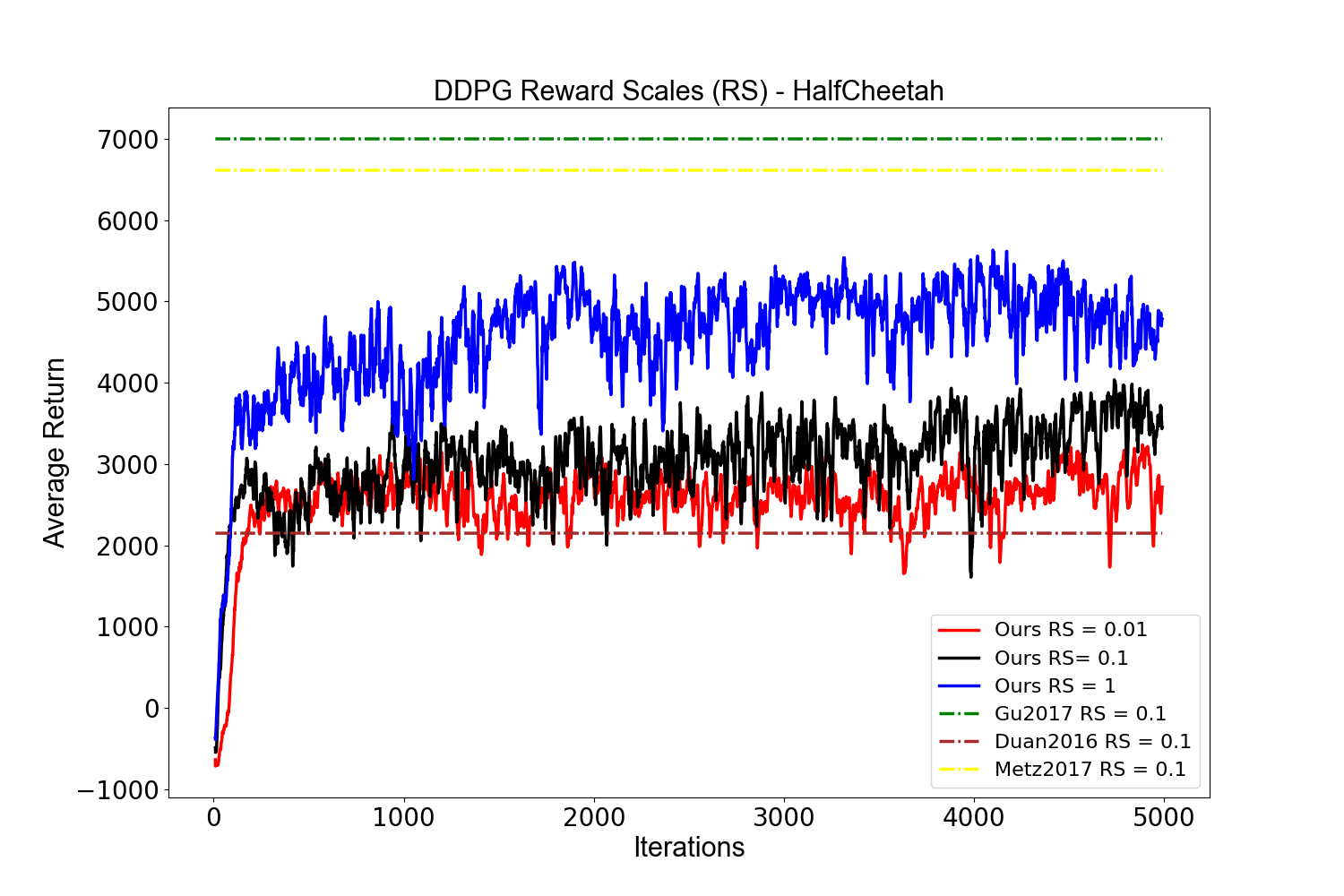}
\includegraphics[width=.5\textwidth]{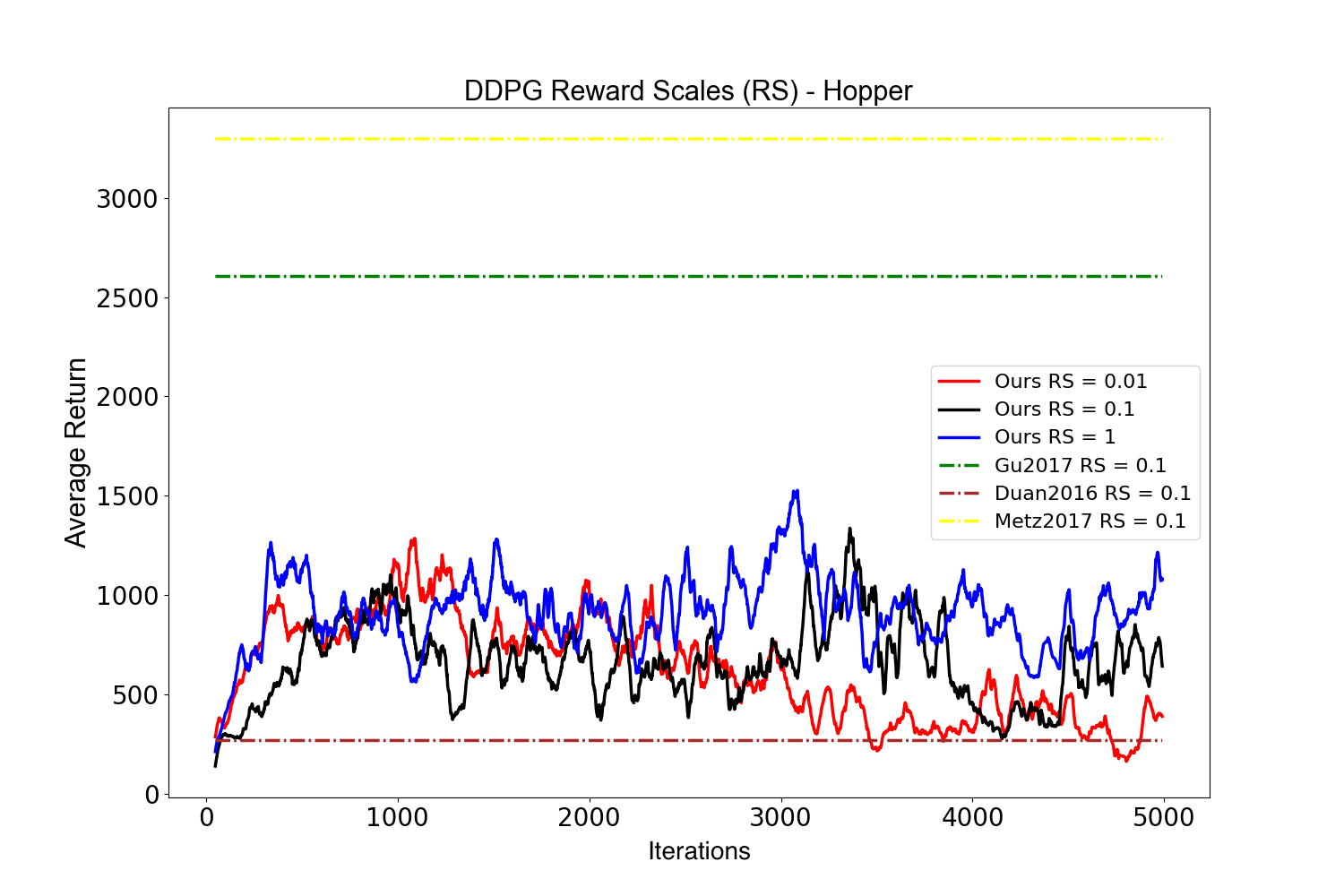}
\caption{DDPG on Half-Cheetah and Hopper - Significance of the Reward Scaling parameter}
\label{ddpg_reward_scale}
\end{figure}

Figure~\ref{ddpg_reward_scale} shows that even though DDPG performance have been reported to be highly susceptible to the reward scaling parameter, our analysis shows that DDPG does not improve by rescaling the rewards. In fact, for the Half-Cheetah environment, we find that no reward scaling (RS=$1$) yields much higher returns, even though \cite{QPROP} and \cite{rllab} have reported an optimal reward scale value to be $0.1$. Furthermore, we highlight that often for DDPG, learning curves are not shown for all environments and only tabular results are presented, making it difficult to compare how reward scaling has affected results in prior work.

\textbf{Actor-Critic Learning Rates : }We further investigate the effects of the actor and critic base learning rates as given in \cite{QPROP} and \cite{rllab}, which both use $0.001, 0.0001$ (for the critic, and actor respectively). Interestingly, we find that the actor and critic learning rates for DDPG have less of an effect on the Hopper environment than the Half-Cheetah environment. This brings into consideration that keeping other parameters fixed, DDPG is not only susceptible to the learning rates, but there are other sources of variation and randomness in the DDPG algorithm.

\begin{figure}[ht!]
\includegraphics[width=.5\textwidth]{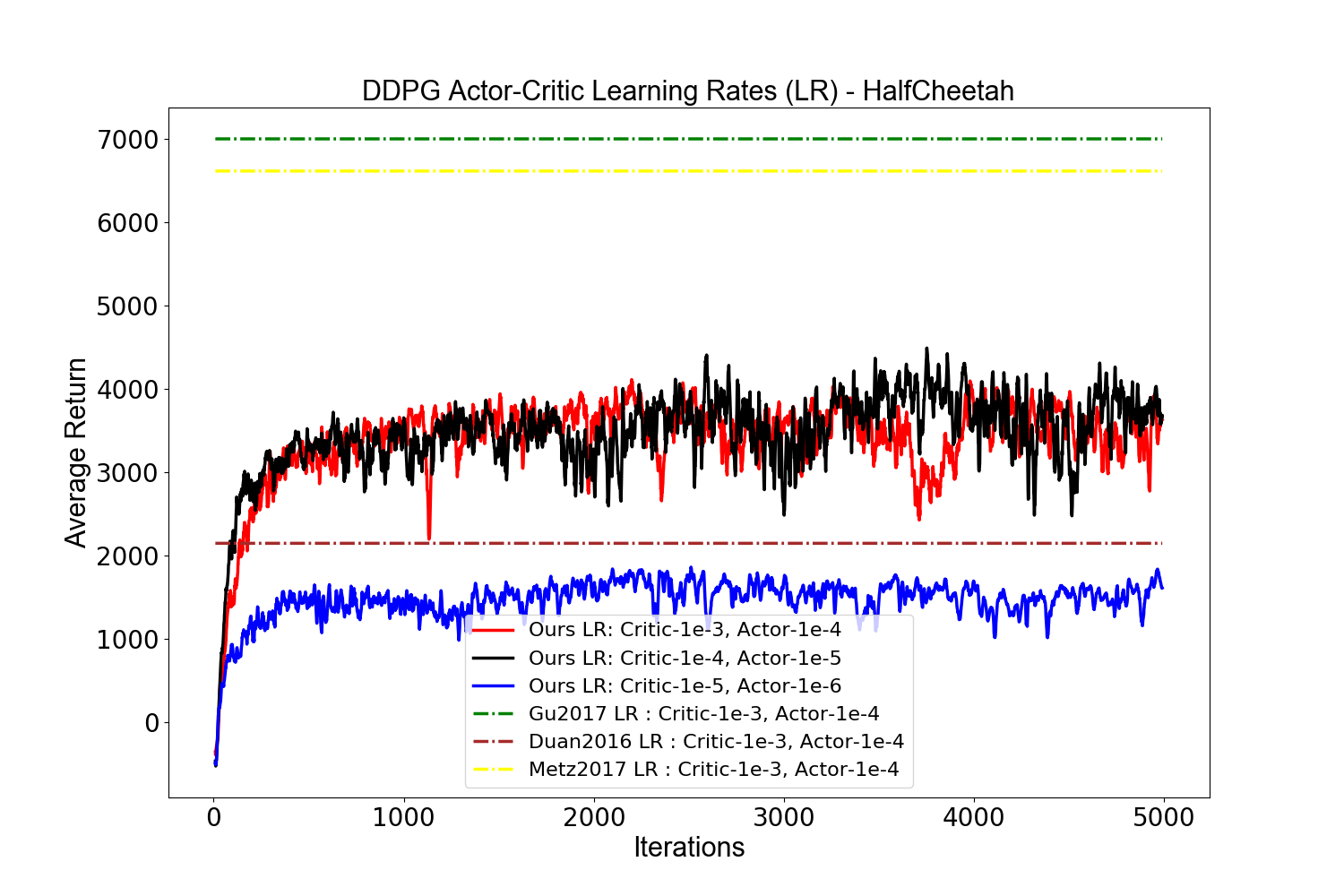}
\includegraphics[width=.5\textwidth]{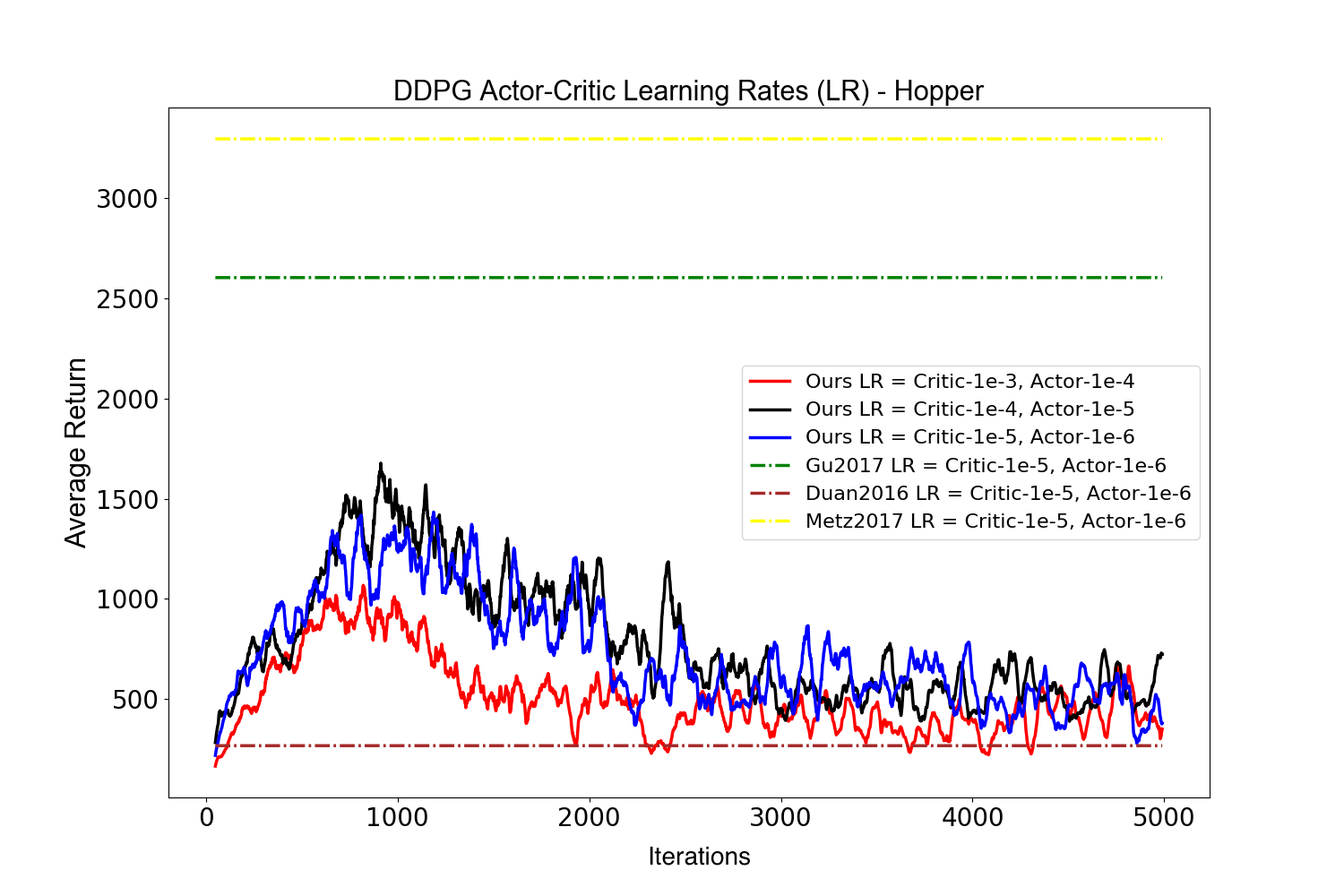}
\caption{DDPG on Half-Cheetah and Hopper - Actor and Critic Learning Rates}
\label{}
\end{figure}

\subsection{General Variance}

We investigate the general variance of multiple trials with different random seeds. Variance across random seeds is of particular interest since it has been noted that in several known codebases, there are implementations\footnote{One such example in the codebase we use here: \url{https://github.com/openai/rllab/blob/master/contrib/rllab\_hyperopt/example/main.py\#L21}.} for searching for the best random seed to use. In particular, we determine  whether it is possible to generate learning curves by randomly averaging trials together (with only the seed varied) such that we see statistically significant differences in the average reward learning curve distributions. Thereby, we wish to determine if it is possible to report significantly worse results on a baseline policy gradient method such as TRPO or DDPG, just by varying the random seed (or significantly better results for the algorithm under investigation by doing so).

\begin{figure}[ht!]
\includegraphics[width=.5\textwidth]{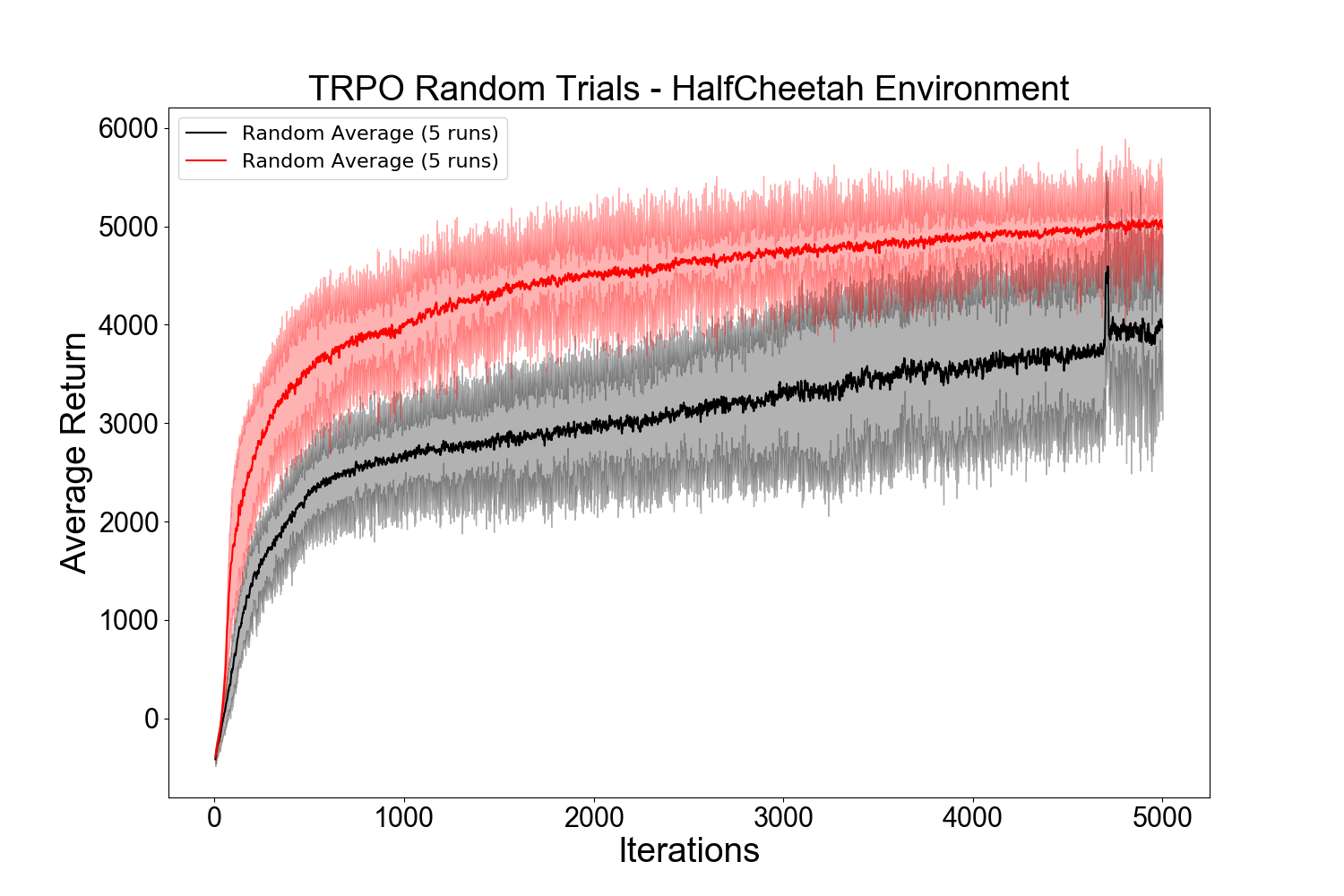}
\includegraphics[width=.5\textwidth]{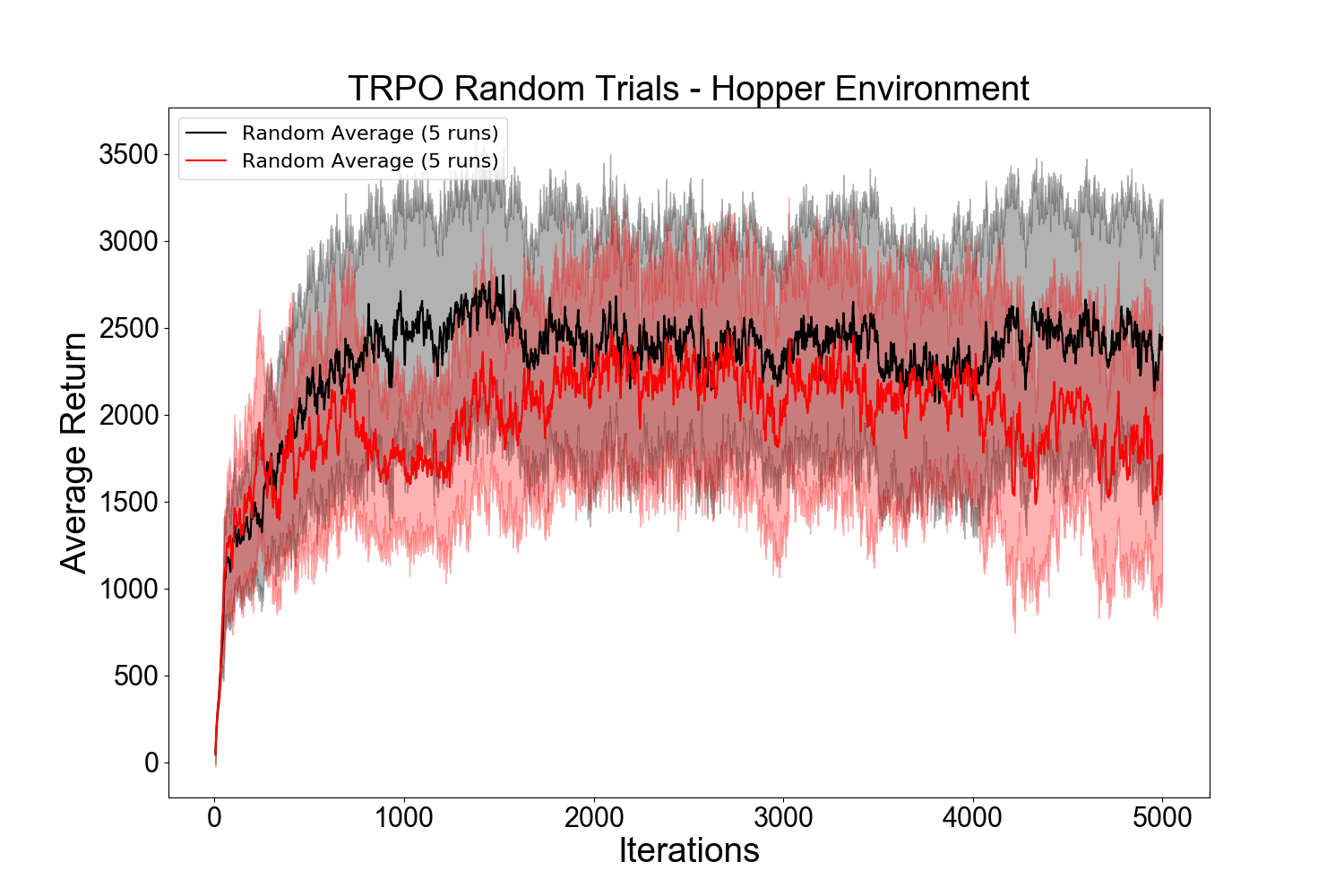}
\caption{TRPO with best hyper-parameter configurations, with average of 5 runs over 2 different set of experiments under same configuration, producing variant results.}
\label{trpo_variance}
\end{figure}

We run a total of $10$ trials with our best tuned hyper-parameter configurations as examined previously. We randomly average two groups of $5$ and plot the results. We find that there can be a significant\footnote{Average 2-sample t-test run across entire training distributions resulting in $t=-9.0916,p=0.0016$ for Half-Cheetah and $t=2.2243,p=0.1825$ for Hopper} difference as seen in Figure~\ref{trpo_variance}. Particularly for Half-Cheetah it is possible to get training curves that do not fall within the same distribution at all, just by averaging different runs with the same hyper-parameters, but random seeds.

\begin{figure}[ht!]
\includegraphics[width=.5\textwidth]{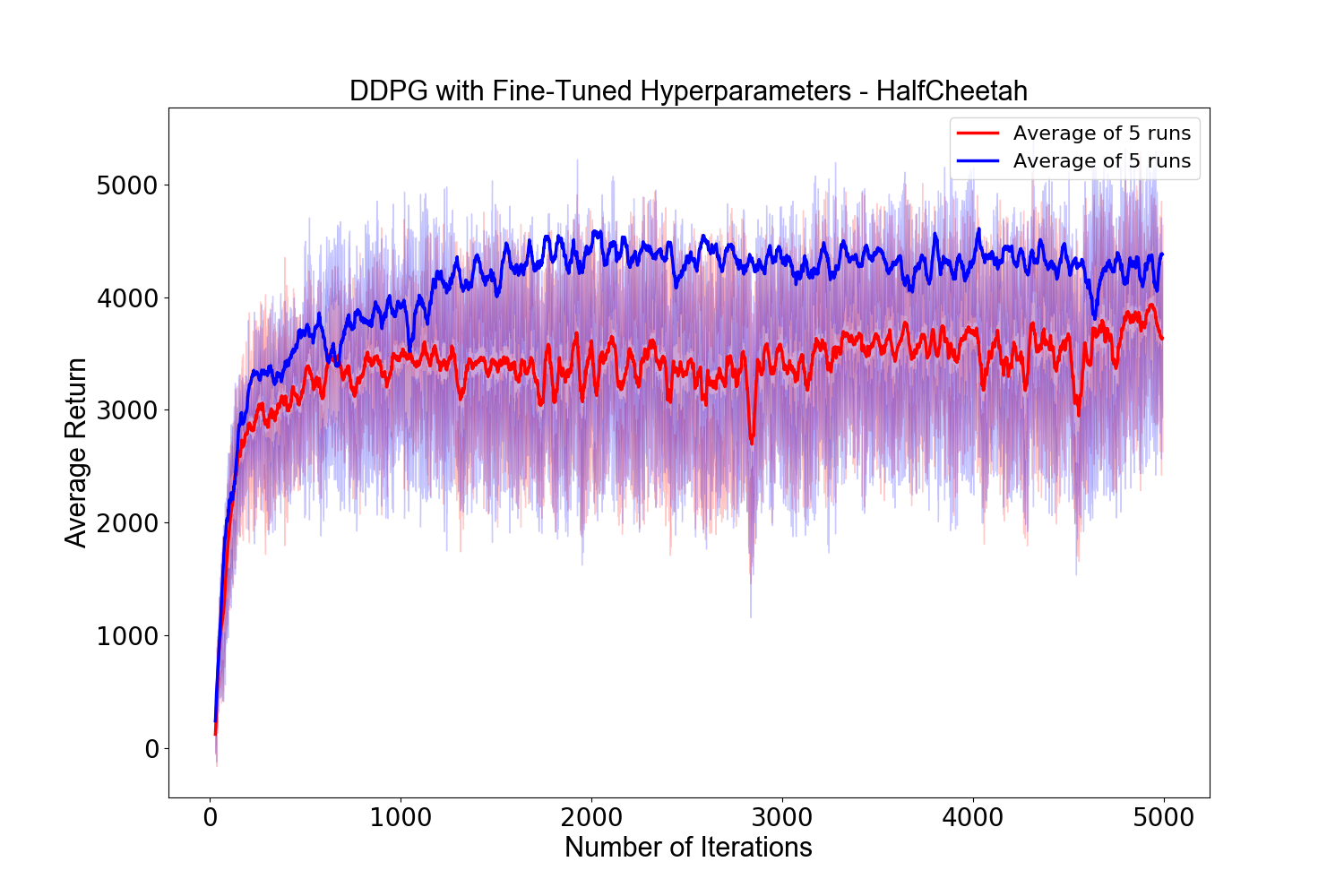}
\includegraphics[width=.5\textwidth]{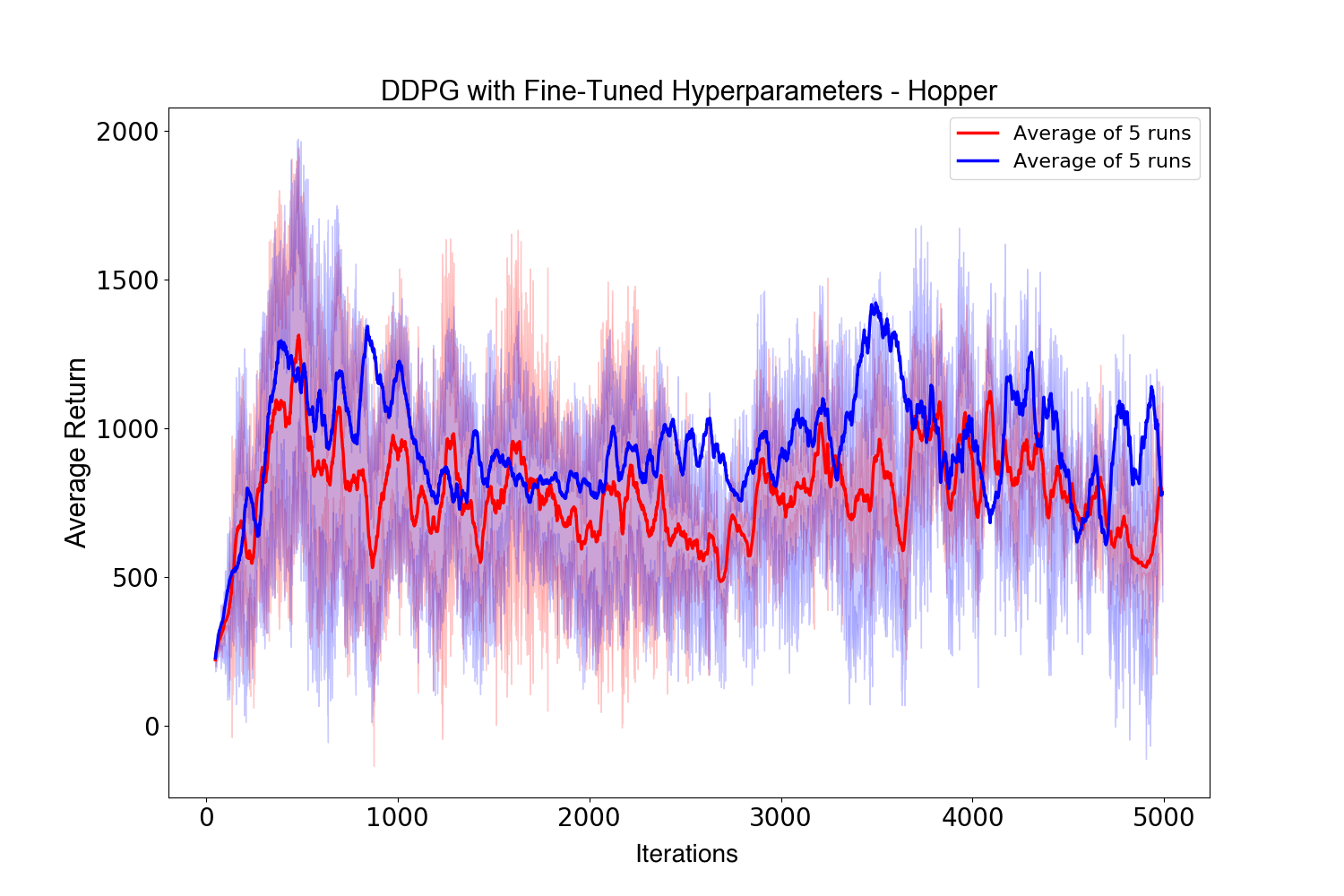}
\caption{DDPG with tuned hyper-parameter configurations, with average of 5 runs over 2 different set of experiments under same configuration, producing variant results.}
\label{ddpg_variance}
\end{figure}

Figure~\ref{ddpg_variance} also shows the significance of DDPG instability. Even with fine-tuned hyper-parameter configurations, our analysis shows that stable results with DDPG, on either of the environments cannot be achieved. This further suggests that there might be randomness due to other external sources which affects performance of DDPG on these continuous control tasks.

Our results show that for both DDPG and TRPO, taking two different average across 5 experiment runs do not necessarily produce the same result, and in fact, there is high variance in the obtained results. This emphasizes the need for averaging many runs together when reporting results using a different randoms seed for each. In this way, future works should attempt to negate the effect of random seeds and environment stochasticity when reporting their results.

\section{Discussion and Conclusion}

Tables~\ref{table:trpo_papers} and~\ref{table:ddpg_papers} highlight results and metrics presented in various related works which compares to TRPO and DDPG (respectively). We include results from an average of 5 runs across the best cross-section of hyper-parameters (based on our previous investigations). We show various metrics at different numbers of iterations such that a fair comparison can be made against reported results from other works. It can be noted that while some works demonstrate similar results to our own, others vary wildly from our own findings. Furthermore, many works only include the Max Average Return, which can be misleading. Due to the variance we have demonstrated here and the difficulty in reproducing these algorithms, it is extremely important for future works to: (1) report all possible metrics to characterize their own algorithms against TRPO and DDPG (particularly Average Return and Standard Deviation of the returns); (2) report all hyper-parameters used for optimization; (3) attempt to use a somewhat optimal set of hyper-parameters; (4) average results on greater than 5 trials and report how many trials are averaged together\footnote{Further investigation needs be done to determine the amount of trials ($N$) necessary to ensure a fair comparison (i.e. for what $N$ would any $N$-sample average always result in a similarly distributed returns, unlike as has been demonstrated to be possible in Figure~\ref{trpo_variance})}. We intend this work to act as both a guide for accomplishing this and a starting point for determining whether observed values are in line with possible best results on Hopper and Half-Cheetah environments for novice researchers in policy gradients.

\begin{table}[htp!]
\tiny\centering\begin{tabular}{|l|l|l|l|l|l|l|l|l|l|}
\hline
Environment & Metric & rllab~\cite{rllab} & QProp~\cite{QPROP} & IPG~\cite{IPG} & TRPO~\cite{TRPO,TRPOGAE}\footnotemark & Ours & Ours & Ours & Ours\\
\hline
\multirow{4}{*}{Half-Cheetah} & Length (iters) & 500 & -- & -- & 500& 500 & 1000 & 2500 & 5000 \\
& Length (episodes)&$\sim$25k&30k&10k&$\sim$12.5k&$\sim$ 12.5k&$\sim$25k&$\sim$62.5k&$\sim$125k\\
 & Average Return & 1914.0 & -- & -- & -- & 3576.08 & 3995.4 & 4638.52 & 5010.83 \\
 & Max Average Return & -- & 4734 & 2889 & 4855.00 & 3980.61 & 4360.77 & 4889.18 & 5197.40 \\
 & Std Return & 120.1 & -- & -- & -- & 434.78 & 502.57 & 419.08 & 443.87 \\
 \hline
\multirow{4}{*}{Hopper} & Length (iters) & 500 & -- & -- & 500& 500 & 1000 & 2500 & 5000 \\
&  Length (episodes)&$\sim$25k&30k&10k&$\sim$22k &$\sim$ 12.5k&$\sim$25k&$\sim$62.5k&$\sim$125k\\
 & Average Return & 1183.3 & -- & -- & -- & 2021.34 & 2285.73 & 2526.41 & 2421.067\\
 & Max Average Return & -- & 2486 & -- & 3668.81 & 3229.14 & 3442.26 & 3456.05 & 3476.00\\
 & Std Return & 150.0 & -- & -- & -- & 654.37 & 757.68 & 714.07 & 796.58 \\
 \hline
\end{tabular}
\caption{Results and descriptions of reported values by various works using TRPO (Hopper and Half-Cheetah environments) as a baseline. "Length(iters)" denotes algorithm iterations and "Length(episodes)" denotes number of episodes.}
\label{table:trpo_papers}
\end{table}

\footnotetext{Results from original implementation evaluation on OpenAI Gym: \url{https://gym.openai.com/evaluations/eval_W27eCzLQBy60FciaSGSJw}; \url{https://gym.openai.com/evaluations/eval_Gudf6XDS2WL76S7wZicLA}}

\begin{table}[htp!]
\tiny\centering\begin{tabular}{|l|l|l|l|l|l|l|l|l|}
\hline
Environment & Metric & rllab~\cite{rllab} & QProp~\cite{QPROP} & SDQN~\cite{SDQN} & Ours & Ours & Ours & Ours\\
\hline
\multirow{4}{*}{Half-Cheetah} & Length (iters) & 500 & $1e^6$ (steps) & -- & 500 & 1000 & 2500 & 5000 \\
& Length (episodes)&$\sim$25k&30k&10k&$\sim$ 12.5k&$\sim$25k&$\sim$62.5k&$\sim$125k\\
 & Average Return &  2148.6  & -- & -- & 2707.1 &  3127.9  & 3547.1 & 3725.3 \\
 & Max Average Return & -- & 7490 & 6614.26 & 3788.2  & 4029.2 & 4460.7  & 4460.7 \\
 & Std Return & -- & -- & -- & 907.1& 784.3  & 634.9 & 512.8  \\
 \hline
\multirow{4}{*}{Hopper} & Length (iters) & 500 & -- & -- & 500 & 1000 & 2500 & 5000 \\
&  Length (episodes)&$\sim$25k&30k&10k&$\sim$ 12.5k&$\sim$25k&$\sim$62.5k&$\sim$125k\\
 & Average Return & 267.1  & -- & -- & 790.6 & 883.6 & 838.7 & 857.6 \\
 & Max Average Return & -- & 2604 & 3296.49 & 1642.1 & 1642.1 & 1642.1 & 1642.1 \\
 & Std Return & -- & -- & -- & 367.9 & 305.2 & 230.9  & 213.7\\
 \hline
\end{tabular}
\caption{Results and descriptions of reported values by various works using DDPG (Hopper and Half-Cheetah environments) as a baseline. "Length(iters)" denotes algorithm iterations and "Length(episodes)" denotes number of episodes.}
\label{table:ddpg_papers}
\end{table}

We present a set of results, highlighting the difficulty in reproducing results with policy gradient methods in reinforcement learning. We show the difficulty of fine-tuning and the significant sources of variance in hyper-parameter selection for both TRPO and DDPG algorithms. Our analysis shows that these state-of-the-art on-policy and off-policy policy gradient methods often suffer from large variations as a result of different hyper-parameter settings. In addition, results across different continuous control domains are not always consistent, as shown in the Hopper and Half-Cheetah experiment results. We find that Half-Cheetah is more susceptible to performance variations from hyper-parameter tuning, while Hopper is not. We posit that this may be due to the difference in stochasticity within the environments themselves. Half-Cheetah has a much more stable dynamics model, and thus is less variant in failure modes. Hopper, on the other hand, is prone to quick failure modes which introduce larger external variance, possibly making tuning difficult.

Based on our experiments, we suggest that the ML research community requires better fine-tuned implementations of these algorithms with provided hyper-parameter presets. These implementations should have benchmark results for a wide range of commonly used tasks. Our analysis shows that due to the under-reporting of hyper-parameters, different works often report different baseline results and performance measures for both TRPO and DDPG. This leads to an unfair comparison of baselines in continuous control environments. Here, we provide some insight into the impact of different hyper-parameters to aid future researchers in finding the ideal baseline configurations.

However, we also suggest that these algorithms are often susceptible to external randomness, introduced by the environment and other external hyper-parameters (e.g reward scale in DDPG) which makes it quite difficult to reproduce results with these state-of-the-art policy gradient algorithms. As such, we provide the aforementioned recommendations in reporting implementation details (provide all hyper-parameters and number of trial experiments), reporting results (report averages and standard deviations, not maximum returns), and implementing proper experimental procedures (average together many trials using different random seeds for each).


\section*{Acknowledgements}
The authors would like to thank Joelle Pineau and David Meger for comments on the paper draft. We would like to thank the McGill University Reasoning and Learning Lab and the Mobile Robotics Lab for allowing an engaging and productive research environment. We would also like to thank Alex Lamb and Anirudh Goyal for providing initial feedback on the direction of this work, as part of the ICML Reproducibility in Machine Learning workshop. This work was supported by the \textit{AWS Cloud Credits for Research} program, McGill Graduate Excellence Scholarships, and \textit{NSERC}.


\begin{thebibliography}{10}

\bibitem{DDPG}
Timothy~P Lillicrap, Jonathan~J Hunt, Alexander Pritzel, Nicolas Heess, Tom
  Erez, Yuval Tassa, David Silver, and Daan Wierstra.
\newblock Continuous control with deep reinforcement learning.
\newblock {\em arXiv preprint arXiv:1509.02971}, 2015.

\bibitem{TRPO}
John Schulman, Sergey Levine, Pieter Abbeel, Michael Jordan, and Philipp
  Moritz.
\newblock Trust region policy optimization.
\newblock In {\em Proceedings of the 32nd International Conference on Machine
  Learning (ICML-15)}, pages 1889--1897, 2015.

\bibitem{TRPOGAE}
John Schulman, Philipp Moritz, Sergey Levine, Michael Jordan, and Pieter
  Abbeel.
\newblock High-dimensional continuous control using generalized advantage
  estimation.
\newblock {\em arXiv preprint arXiv:1506.02438}, 2015.

\bibitem{DQN}
Volodymyr Mnih, Koray Kavukcuoglu, David Silver, Andrei~A. Rusu, Joel Veness,
  Marc~G. Bellemare, Alex Graves, Martin Riedmiller, Andreas~K. Fidjeland,
  Georg Ostrovski, Stig Petersen, Charles Beattie, Amir Sadik, Ioannis
  Antonoglou, Helen King, Dharshan Kumaran, Daan Wierstra, Shane Legg, and
  Demis Hassabis.
\newblock Human-level control through deep reinforcement learning.
\newblock {\em Nature}, 518(7540):529--533, 02 2015.

\bibitem{levine}
Sergey Levine, Chelsea Finn, Trevor Darrell, and Pieter Abbeel.
\newblock End-to-end training of deep visuomotor policies.
\newblock {\em CoRR}, abs/1504.00702, 2015.

\bibitem{Gu-Model-Based}
Shixiang Gu, Timothy~P. Lillicrap, Ilya Sutskever, and Sergey Levine.
\newblock Continuous deep q-learning with model-based acceleration.
\newblock In {\em Proceedings of the 33nd International Conference on Machine
  Learning, {ICML} 2016, New York City, NY, USA, June 19-24, 2016}, pages
  2829--2838, 2016.

\bibitem{MuJoCo}
Emanuel Todorov, Tom Erez, and Yuval Tassa.
\newblock Mujoco: {A} physics engine for model-based control.
\newblock In {\em 2012 {IEEE/RSJ} International Conference on Intelligent
  Robots and Systems, {IROS} 2012, Vilamoura, Algarve, Portugal, October 7-12,
  2012}, pages 5026--5033, 2012.

\bibitem{Gym}
Greg Brockman, Vicki Cheung, Ludwig Pettersson, Jonas Schneider, John Schulman,
  Jie Tang, and Wojciech Zaremba.
\newblock Openai gym.
\newblock {\em CoRR}, abs/1606.01540, 2016.

\bibitem{rllab}
Yan Duan, Xi~Chen, Rein Houthooft, John Schulman, and Pieter Abbeel.
\newblock Benchmarking deep reinforcement learning for continuous control.
\newblock In {\em Proceedings of the 33rd International Conference on Machine
  Learning (ICML)}, 2016.

\bibitem{QPROP}
Shixiang Gu, Timothy Lillicrap, Zoubin Ghahramani, Richard~E Turner, and Sergey
  Levine.
\newblock Q-prop: Sample-efficient policy gradient with an off-policy critic.
\newblock {\em arXiv preprint arXiv:1611.02247}, 2016.

\bibitem{SDQN}
Luke Metz, Julian Ibarz, Navdeep Jaitly, and James Davidson.
\newblock Discrete sequential prediction of continuous actions for deep rl.
\newblock {\em arXiv preprint arXiv:1705.05035}, 2017.

\bibitem{IPG}
Shixiang Gu, Timothy Lillicrap, Zoubin Ghahramani, Richard~E Turner, Bernhard
  Sch{\"o}lkopf, and Sergey Levine.
\newblock Interpolated policy gradient: Merging on-policy and off-policy
  gradient estimation for deep reinforcement learning.
\newblock {\em arXiv preprint arXiv:1706.00387}, 2017.

\bibitem{houthooft2016vime}
Rein Houthooft, Xi~Chen, Yan Duan, John Schulman, Filip De~Turck, and Pieter
  Abbeel.
\newblock Vime: Variational information maximizing exploration.
\newblock In {\em Advances in Neural Information Processing Systems}, pages
  1109--1117, 2016.

\bibitem{rajeswaran2016epopt}
Aravind Rajeswaran, Sarvjeet Ghotra, Sergey Levine, and Balaraman Ravindran.
\newblock Epopt: Learning robust neural network policies using model ensembles.
\newblock {\em arXiv preprint arXiv:1610.01283}, 2016.

\end{thebibliography}

\end{document}